\documentclass{article}

\PassOptionsToPackage{numbers}{natbib}

\usepackage[preprint]{neurips_2025}

\usepackage[utf8]{inputenc} 
\usepackage[T1]{fontenc}    
\usepackage[colorlinks=true, citecolor=blue, linkcolor=black, urlcolor=Violet]{hyperref}       
\usepackage{url}            
\usepackage{booktabs}       
\usepackage{amsfonts}       
\usepackage{nicefrac}       
\usepackage{microtype}      
\usepackage[dvipsnames]{xcolor}         
\usepackage{subfigure}
\usepackage{multirow}
\usepackage{multicol} 
\usepackage{pifont}
\usepackage{graphicx}
\usepackage{algorithm}
\usepackage{algpseudocode}
\usepackage{bold-extra}
\usepackage{caption}
\usepackage{amsmath}
\usepackage{amssymb}
\usepackage[font={footnotesize}]{caption}
\usepackage{subcaption}

\usepackage[utf8]{inputenc} 
\usepackage[T1]{fontenc}    
\usepackage{url}            
\usepackage{booktabs}       
\usepackage{amsfonts}       
\usepackage{nicefrac}       
\usepackage{microtype}      

 \usepackage{graphicx}
\usepackage{subfigure} 
\usepackage{pifont}
\usepackage{adjustbox}
\usepackage{lipsum}
\usepackage{color}
\usepackage{wrapfig}
\usepackage{booktabs}
\usepackage[textsize=scriptsize]{todonotes}
\usepackage{multirow,mathtools}
\usepackage{threeparttable}
\usepackage{tikz-cd}

\definecolor{lightblue}{rgb}{0.68, 0.85, 0.9}
\definecolor{lightgreen}{rgb}{0.56, 0.93, 0.56}
\definecolor{lightskyblue}{rgb}{0.53, 0.81, 0.98}
\definecolor{non-photoblue}{rgb}{0.64, 0.87, 0.93}
\definecolor{magicmint}{rgb}{0.67, 0.94, 0.82}
\definecolor{mossgreen}{rgb}{0.68, 0.87, 0.68}
\definecolor{salmon}{rgb}{1.0, 0.55, 0.41}
\definecolor{babypink}{rgb}{0.96, 0.76, 0.76}
\definecolor{darkgreen}{rgb}{0, 0.7, 0}

\DeclareMathOperator*{\minimize}{\text{minimize}}

\DeclareMathAlphabet\mathbfcal{OMS}{cmsy}{b}{n}

\usepackage{color, colortbl}
\definecolor{Gray}{gray}{0.93}
\definecolor{Orange}{rgb}{1,0.5,0}
\definecolor{DGray}{gray}{0.83}
\definecolor{LightCyan}{rgb}{0.88,1,1}

\usepackage{dsfont}
\usepackage{enumerate}
\usepackage{graphicx}
\usepackage{threeparttable}

\definecolor{Red}{rgb}{0.6,0,0}
\definecolor{Blue}{rgb}{0,0,0.8}
\definecolor{Green}{rgb}{0,0.6,0.9}
\definecolor{airforceblue}{rgb}{0.36, 0.54, 0.66}
\definecolor{ao(english)}{rgb}{0.0, 0.5, 0.0}
\definecolor{azure(colorwheel)}{rgb}{0.0, 0.5, 1.0}
\definecolor{crimson}{rgb}{0.86, 0.08, 0.24}
\definecolor{darkcerulean}{rgb}{0.03, 0.27, 0.49}
\definecolor{cobalt}{rgb}{0.0, 0.28, 0.67}
\definecolor{rosegold}{rgb}{0.72, 0.43, 0.47}
\definecolor{orange-red}{rgb}{1.0, 0.27, 0.0}
\definecolor{mountainmeadow}{rgb}{0.19, 0.73, 0.56}
\definecolor{malachite}{rgb}{0.04, 0.85, 0.32}
\definecolor{darkblue}{rgb}{0.0, 0.0, 0.55}
\definecolor{customred}{rgb}{1, 0.85, 0.85}
\definecolor{customcitecolor}{rgb}{0.7, 0.5, 1}
\definecolor{custompink}{rgb}{0.8, 0.3, 0.3}
\definecolor{customgreen}{rgb}{0.3, 0.8, 0.3}
\definecolor{Lightgreen}{rgb}{0.8, 1, 0.9}
\definecolor{LightCyan}{rgb}{0.8, 0.9, 1}
\definecolor{mygray}{gray}{0.8}

\usepackage[most]{tcolorbox}
\newtcolorbox{mybox}[2][]{%
  attach boxed title to top center
               = {yshift=-8pt},
  colback      = Gray,
  colframe     = black,
  fonttitle    = \bfseries,
  colbacktitle = white,
  title        = #2,#1,
  enhanced,
}

\DeclarePairedDelimiterX{\inp}[2]{\langle}{\rangle}{#1, #2}

\makeatletter
\newcommand*{\rom}[1]{\expandafter\@slowromancap\romannumeral #1@}
\makeatother
\newcommand{\mycomment}[1]{}

\usepackage{pifont}

\usepackage{color, colortbl}
\definecolor{Gray}{gray}{0.93}
\definecolor{Orange}{rgb}{1,0.5,0}
\definecolor{DGray}{gray}{0.83}
\definecolor{LightCyan}{rgb}{0.88,1,1}

\definecolor{WarnREd}{rgb}{1,0.4,0.4}
\definecolor{WarnOrange}{rgb}{1,0.682,0.502}
\definecolor{WarnPink}{rgb}{0.9176, 0.7215, 0.7215}
\definecolor{GoodGreen}{rgb}{0.5019, 0.9215, 0.6039}
\definecolor{deepgreen}{RGB}{0,128,0}   
\definecolor{ochreyellow}{RGB}{204,119,34} 
\definecolor{GoodGreen}{rgb}{0.5019, 0.9215, 0.6039}
\definecolor{NiceYellow}{rgb}{0.98, 0.92, 0.60}

\newcommand{\congrat}[1]{\sethlcolor{GoodGreen}\hl{#1}}

\newcommand{\warn}[1]{\sethlcolor{WarnPink}\hl{#1}}

\newcommand{\warncellred}{\cellcolor{WarnPink}}

\newcommand{\secbest}[1]{\sethlcolor{NiceYellow}\hl{#1}}

\usepackage[T1]{fontenc}

\definecolor{styleblue}{HTML}{504099}
\definecolor{mypurple}{HTML}{9391ff}
\usepackage{wrapfig}

\usepackage{multirow,mathtools }

\usepackage{pifont}
\usepackage{color, colortbl}

\usepackage{blindtext}
\usepackage{lipsum}

\usepackage{multirow}
\usepackage{listings}

\usepackage{bbm}

\usepackage [english]{babel}
\usepackage [autostyle, english = american]{csquotes}
\usepackage[nottoc]{tocbibind}

\usepackage{pifont}
\usepackage{url}
\usepackage[most]{tcolorbox}

\usepackage{lipsum}
\usepackage{soul}
\usepackage{xcolor}
\usepackage{wrapfig}
\usepackage{multirow,mathtools } 

\usepackage{adjustbox}
\MakeOuterQuote{"}

\usepackage{microtype}
\usepackage{graphicx}
\usepackage{booktabs} 

\usepackage{amsmath}
\usepackage[capitalize,noabbrev]{cleveref}
\usepackage{tablefootnote}

\definecolor{ceruleanblue}{rgb}{0.16, 0.32, 0.75}

\hypersetup{
 colorlinks=true,
 citecolor=ceruleanblue,
 linkcolor=ceruleanblue,
 urlcolor=black}

\usepackage{amsmath,amsfonts,bm,nicefrac}

\def\eqref#1{Eq.~\ref{#1}}

\def\1{\bm{1}}

\def\vx{{\bm{x}}}

\DeclareMathAlphabet{\mathsfit}{\encodingdefault}{\sfdefault}{m}{sl}
\SetMathAlphabet{\mathsfit}{bold}{\encodingdefault}{\sfdefault}{bx}{n}

\newcommand{\KL}{D_{\mathrm{KL}}}

\newcommand{\snr}{\alpha}

\newcommand{\logsnr}{\alpha}
\newcommand{\half}{\nicefrac{1}{2}}

\def\eps{{\bm \epsilon}}

\newcommand{\be}{\begin{eqnarray} \begin{aligned}}
\newcommand{\ee}{\end{aligned} \end{eqnarray} }
\newcommand{\benn}{\begin{eqnarray*} \begin{aligned}}
\newcommand{\eenn}{\end{aligned} \end{eqnarray*} }
\newcommand{\xhat}{\hat{\bm x}}

\newcommand{\ds}{\frac{d}{d \alpha}}

\newcommand{\vz}{\bm x_\logsnr}

\DeclareMathOperator{\mmse}{mmse}

\newcommand{\norm}[1]{{\| #1 \|^2 }}

\usepackage{enumitem}

\title{
Compensation-free Machine Unlearning \\ in Text-to-Image Diffusion Models \\ by Eliminating the Mutual Information}

\author{
Xinwen Cheng, Jingyuan Zhang, Zhehao Huang, Yingwen Wu, Xiaolin Huang\thanks{Corresponding Author.} \\
\\
Institute of Image Processing and Pattern Recognition\\
School of Automation and Intelligent Sensing\\
Shanghai Jiao Tong University\\
\texttt{\{xinwencheng,tonyzhang666,kinght\_h,yingwenwu,xiaolinhuang\}@sjtu.edu.cn}
}

\begin{document}

\maketitle
   \begin{abstract}
The powerful generative capabilities of diffusion models have raised growing privacy and safety concerns regarding generating sensitive or undesired content. In response, machine unlearning (MU) -- commonly referred to as concept erasure (CE) in diffusion models -- has been introduced to remove specific knowledge from model parameters meanwhile preserving innocent knowledge. Despite recent advancements, existing unlearning methods often suffer from excessive and indiscriminate removal, which leads to substantial degradation in the quality of innocent generations. To preserve model utility, prior works rely on \textit{compensation}, \textit{i.e.}, re-assimilating a subset of the remaining data or explicitly constraining the divergence from the pre-trained model on remaining concepts. However, we reveal that generations beyond the compensation scope still suffer, suggesting such post-remedial compensations are inherently insufficient for preserving the general utility of large-scale generative models. Therefore, in this paper, we advocate for developing \textit{compensation-free} concept erasure operations, which precisely identify and eliminate the undesired knowledge such that the impact on other generations is minimal. In technique, we propose to \textbf{MiM-MU}, which is to unlearn a concept by minimizing the mutual information with a delicate design for computational effectiveness and for maintaining sampling distribution for other concepts. Extensive evaluations demonstrate that our proposed method achieves effective concept removal meanwhile maintaining high-quality generations for other concepts, and remarkably, without relying on any post-remedial compensation for the first time. 

\end{abstract}

\begin{figure}[htbp]
  \centering
  \includegraphics[width=0.99\linewidth]{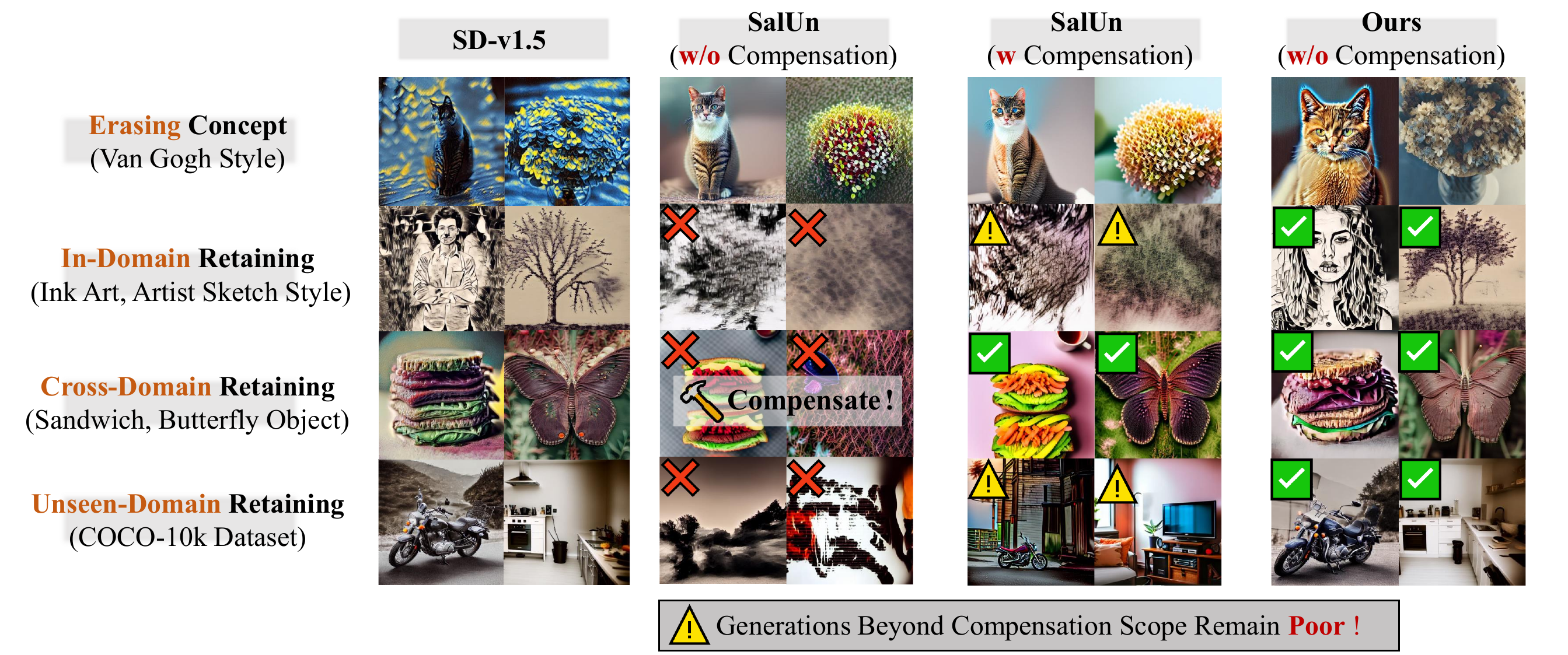}
\vspace*{-1mm}
  \caption{\textbf{Retainability across different concept domains of SalUn~\cite{fan2023salun} (compensation-dependent) and our method (comepnsation-free) when unlearning ``Van Gogh'' style, revealing the failure of post-remedial compensation to restore generations beyond explicitly compensated concepts.} Images in the first three rows are generated with concepts in UnlearnCanvas benchmark (which is elaborated in Sec.~\ref{sec:exp_setup}) in the form of \textit{"A \{object\} in \{artist\} style"} and images in the last row are respectively generated with prompts from COCO-10k dataset \textit{"The shiny motorcycle has been put on display"} and \textit{"A kitchen filled with furniture and a stove top oven"}. SalUn with compensation refers to compensating concepts in the cross-domain of UnlearnCanvas (\textit{i.e.}, 20 objects including ``Sandwich'' and ``Butterfly'') by re-assimilating corresponding data. However, as can be seen, such compensation failed to restore concepts beyond the compensated scope, \textit{i.e.}, ``Ink Art'' and ``Artist Sketch'' style in the same style domain as ``Van Gogh'' and ``kitch'' and ``stove'' in the COCO-10k dataset. }
\label{fig:introduction_ood}
\vspace*{-4mm}
\end{figure}

\section{Introduction}
Diffusion models (DM) have made remarkable strides in recent years, showing a great ability to generate realistic images. However, this powerful generative capability raises pressing privacy and safety concerns, particularly regarding the potential generation of undesirable content, such as Not Safe For Work (NSFW) images \cite{fan2023salun,rando2022red,schramowski2023safe}, copyright-infringing pictures \cite{shan2023glaze, wang2023diagnosis, an2024rethinking} and training data replication \cite{somepalli2023understanding, carlini2023extracting}. To effectively and thoroughly disable these generations without the need to re-train the model, \textit{``machine unlearning''} (MU), which is also known as \textit{``concept erasing''} (CE) for diffusion models, has emerged as a critical approach. MU aims at rapidly and seamlessly removing concept-related information from model parameters meanwhile maintaining the model performance on other generations.

A variety of competitive methodologies have been developed to manipulate the unlearned model's behavior on the erasing concept to unlearn. The manipulations can be broadly categorized into three types: (1) \textbf{Retargeting} model outputs of erasing concept to that of an anchor concept~\cite{fan2023salun, kumari2023ablating,gandikota2024unified, heng2024selective, lu2024mace}, (2) \textbf{Repelling} from the pre-trained model's behavior~\cite{gandikota2023erasing,zhao2024separable} and (3) \textbf{Suppressing} intermediate activations (\textit{e.g.}, the cross-attention maps) to reduce responses~\cite{zhang2024forget,li2024get, wu2024scissorhands}. However, recent evaluations on the comprehensive  UnlearnCanvas benchmark~\cite{zhang2024unlearncanvas} have revealed that above erasures significantly degrade innocent generations due to their indiscriminate and aggressive removal~\cite{kumari2023ablating,fan2023salun,gandikota2023erasing,lu2024mace,zhao2024separable}, necessitating additional maintenance on the remaining data to preserve model utility.


Although such compensations achieve some effectiveness and are widely adopted and allowed in previous works, we raise the attention that their power is not as satisfactory as desired. Prior work typically assesses the retainability of the unlearned model within the same scope as the compensated concepts. However, in \textbf{Fig.}\ref{fig:introduction_ood}, we illustrate the retainability of our method and SalUn~\cite{fan2023salun} (which achieves the best performance in UnlearnCanvas benchmark~\cite{zhang2024unlearncanvas}) across different concept domains, revealing that generations regarding concepts \textit{outside} the explicitly maintained scope remain significantly degraded. Specifically, we highlight two \textbf{fundamental limitations} inherent to post-remedial compensations: (1) The inadvertent damages introduced by unlearning are usually difficult to diagnose, potentially creating subtle but cumulative performance degradation; (2) Compensation is typically restricted to a narrowly presupposed scope, whereas generative models are expected to handle a vast and diverse range of concepts so that generations beyond compensation might remain poor. These underscore the urgent need for developing \textit{compensation-free} unlearning approaches for large-scale generative models, which could effectively eliminate undesired concepts meanwhile minimally impacting other generations when erasing.

Since there is no chance to relearn, a compensation-free unlearning necessitates precise identification and removal of target knowledge.  
To achieve this goal, we propose to minimize
$$p(y|x), \forall x\sim\mathcal{S}_{\theta_U},$$ 
where $y$ denotes the erasing concept and $\mathcal{S}_{\theta_U}$ denotes the sampling distribution of the unlearned model $\theta_U$. The idea is that when 
$p(y|x)\xrightarrow{}0$, the generations by the unlearned model are devoid of any semantic associated with the erasing concept $y$ and thus the probability of being identified as $y$ is very small. By Bayes' rule, minimizing $p(y|x)$ is equivalent to minimizing the likelihood ratio $p(x|y)/p(x)$, for $p(y)$ is a constant independent of the generated $x$. This likelihood ratio quantifies \emph{mutual information} between the textual concept $y$ and the generated image $x$, \textit{i.e.}, $\mathcal{I}(x,y)=\log p(x|y)-\log p(x)$. Our method is hence named Mutual Information Minimization \textbf{(MiM-MU)}.

As theoretically elucidated by \citet{kong2023information}, a pre-trained diffusion model allows an \textit{exact} density estimation for $p(x)$ and $p(x|y)$. This enables MiM-MU to leverage the pre-trained diffusion model as a competitive discriminator to quantify concept-related information in images generated by the unlearned model, followed by back-propagating to the unlearned model to minimize such mutual information. Beyond the core principle, we also address two critical technical issues: (1) \textit{Efficiency:} to facilitate the optimization, we analyze the back-propogated gradient flow and identify that the Jacobian of the pre-trained model could be reasonably omitted to reduce computational overhead; (2) \textit{Minimal Interference:} ensure the unlearning minimally affects innocent generations, we propose that the unlearned sampling distribution should remain as close as possible to that of the pre-trained model while minimizing mutual information and identify the one as the marginal distribution of the pre-trained model. We evaluate style and object unlearning on a comprehensive benchmark UnlearnCanvas~\cite{zhang2024unlearncanvas}, including 50 styles and 20 objects. Remarkably, our unlearning method achieves favorable concept erasure meanwhile well preserving the general utility without relying on any post-remedial compensation for the first time.

We summarize our key contributions as follows:

\noindent\ding{182} We provide a principled formulation of the concept of erasure objective in diffusion models from an information-theoretic perspective, by quantifying the mutual information between textual concepts and unlearned sampling distribution with the pre-trained diffusion model.

\noindent\ding{183} To preserve the model’s general utility during unlearning, we propose to align the sampling distribution of the unlearned model with the marginal distribution of the pre-trained model, which is identified as the closest concept-irrelevant distribution to the original. 

\noindent\ding{184} We reveal that existing post-remedial compensation strategies exhibit limited recovery
and fail for generations beyond the compensation scope, advocating for more benign erasure rather than excessive removal and unreliable compensations. In contrast, our method achieves faithful concept erasure while preserving the general model utility without any compensation for the first time.

\vspace{-2mm}
\section{Related Work}
\vspace{-2mm}
\noindent\textbf{Machine unlearning for Text-to-image (T2I) Diffusion Models.} The powerful generation capabilities of T2I models also bring about safety and privacy concerns of undesirable contents, including  NSFW generations \cite{rando2022red,schramowski2023safe}, artistic copyright \cite{shan2023glaze, wang2023diagnosis, an2024rethinking}, and training data replication \cite{carlini2023extracting, somepalli2023understanding}. While certain undesired generations can be mitigated through filtering mechanisms or modified inference-time guidance \cite{schramowski2023safe, brack2023sega, li2024get, yoon2024safree, jain2024trasce}, these strategies are easily circumvented. Machine Unlearning (MU), also referred to as concept erasing (CE) in diffusion models, offers a more thorough solution by permanently removing undesirable knowledge from the model’s parameters while preserving knowledge unrelated to the target concept.
Existing erasing ideas can be broadly categorized into 3 types: retargeting, repelling, and suppressing. Retargeting retargets model outputs on the erasing concept to that of a neighborhood anchor concept \cite{kumari2023ablating,kim2023towards, gandikota2024unified, heng2024selective, lu2024mace, fan2023salun}, which should be sufficiently distinct from the erasing concept to achieve effective unlearning, meanwhile not too divergent to damage benign generations severely, requiring delicate design and inspection to trade-off the unlearning and maintaining.  Re-pelling steers the classifier-free-guidance (CFG) term away from the original denoising trajectory, thereby avoiding the search for a suitable anchor concept \cite{gandikota2023erasing,zhao2024separable}.  Suppressing aims to locate the erasing concept associated knowledge and obliviate them on purpose, mainly focusing on diminishing activations of the cross-attention map \cite{zhang2024forget,li2024get, wu2024scissorhands}.

\begin{wraptable}{r}{0.50\textwidth}
    \centering
    \vspace*{-4mm}
    \caption{The constraints and manipulated objective to maintain the performance on the remaining concepts. }
    \label{tab:maintenance_method}
    \vspace*{-1mm}
    \resizebox{0.50\textwidth}{!}{ \begin{tabular}{c|c|c}
        \toprule[1pt]
        \midrule
        \textbf{Manner} & \textbf{Objective} & \textbf{Method}  \\
        \midrule
        None & \cellcolor{Gray} - & \cellcolor{Gray} ESD \cite{gandikota2023erasing}, FMN \cite{zhang2024forget}, SDD \cite{kim2023towards} \\
        \midrule
        \multirow{2}{*}{\begin{tabular}[c]{@{}c@{}}Reduce Interference  \\ to Enhance Locality\\ \end{tabular}} & \begin{tabular}[c]{@{}c@{}}Gradient\end{tabular}  &  DoCo\cite{wu2024unlearning}  \\
        \cline{2-3}
         & \cellcolor{Gray} \begin{tabular}[c]{@{}c@{}}Prediction\end{tabular} & \cellcolor{Gray} SepME \cite{zhao2024separable}\\
        \midrule
        \multirow{3}{*}{\begin{tabular}[c]{@{}c@{}} Encourage Alignment  \\ to Repair Performance\\\end{tabular}} & \begin{tabular}[c]{@{}c@{}} Attention\\Map\end{tabular}  & SEOT \cite{li2024get} \\ 
        \cline{2-3}
         & \cellcolor{Gray} \begin{tabular}[c]{@{}c@{}}Text\\Embedding\end{tabular} &\cellcolor{Gray} \begin{tabular}{@{}c@{}} CA \cite{kumari2023ablating}, UCE \cite{gandikota2024unified},\\REFACT\cite{arad2023refact},  MACE \cite{lu2024mace}, RECE \cite{gong2024reliable}\end{tabular}
         \\[0.2em]
        \cline{2-3}
       & \begin{tabular}[c]{@{}c@{}}Predicted\\Noise\end{tabular} & \begin{tabular}{@{}c@{}} 
    SalUn\cite{fan2023salun}, EDiff \cite{wu2024erasediff},\\  
    SHS \cite{wu2024scissorhands}, AdvUnlearn \cite{zhang2024defensive},  
    SafeGen\cite{li2024safegen}, 
\end{tabular} \\
        \midrule
        \bottomrule[1pt]
    \end{tabular}}
    \vspace*{-4mm}
    \label{tab:repair_work}
\end{wraptable}
\noindent\textbf{Model utility maintenance in current T2I MU.} Since there are usually multiple concepts in an image, erasing inadvertently affects unrelated concepts, leading to significant degradation in model utility. Current work either reduces interference with the remaining concepts when unlearning or compensates for damage after unlearning to preserve model utility. The former enhances the locality of the forgetting operation ~\cite{zhao2024separable, wu2024unlearning}, and the latter constrains the divergence before and after unlearning on the remaining concepts~\cite{kumari2023ablating,gandikota2024unified,arad2023refact,li2024get, lu2024mace,gong2024reliable}. \textbf{Tab.}\ref{tab:maintenance_method} summarizes common maintenance measures. However, even with explicit maintenance, most of the existing methods still fail to preserve the model utility, as revealed by a comprehensive benchmark UnlearnCanvas~\cite{zhang2024unlearncanvas}. In this paper, we raise the attention that the additional compensation is limited and insufficient for a large-scale generative task, and a practical concept erasing method should prioritize a non-interfering erasure over excessive and indiscriminate removal.

\section{MUMI: Machine Unlearning by Eliminating the Mutual Informatoin}\label{sec:method}

\subsection{Problem Statement}
MU in T2I diffusion models (more detailed preliminaries can be found in Appendix~\ref{sec:preliminary}) aims to prevent generating any erasing concept-related content meanwhile preserving model utility on other generations. A thorough erasure demands that any generation $\vx$ by the unlearned model should not contain any semantics of the erasing concept. Mathematically, the probability of $\vx$ being classified into the erasing concept $y$ by an oracle classifier approaches 0, \textit{i.e.}, $\forall \vx\sim\mathcal{S}_{\theta_U},\,p(y|\vx)\xrightarrow{}0$ where $\mathcal{S}_{\theta_{\text{U}}}$ denotes the sampling distribution of the unlearned model $\theta_U$: 
{\small{\begin{equation}
    \minimize_{\theta_U}\;\mathcal{D}_{\text{forget}} := \mathbb{E}_{\vx\sim \mathcal{S}_{\theta_U}} \left[\log p(y|\vx)\right].
    \label{eq:forget}
\end{equation}}}There are two main challenges when optimizing \eqref{eq:forget}: (1) $p(y|\vx)$ requires an oracle classifier to identify the quantity of the erasing concept $y$ related semantics in generated image $\vx$; (2) Directly back-propogating \eqref{eq:forget} to optimize the whole sampling distribution of the unlearned model $\mathcal S_{\theta_U}$ might be computationally expensive and impractical.

In this section, we first show that diminishing $p(y|\vx)$ is to diminish the mutual information between textual concept $y$ and corresponding image $\vx$, \textit{i.e.}, $\log p(\vx|y) - \log p(\vx)$, which could be well estimated by the pre-trained diffusion model from an information-theoretic perspective. Instead of straightforwardly optimizing the sampling distribution of the unlearned model through gradient descent, we characterize the family of distributions that have minimal mutual information and propose to approach the one that is \textit{closest} to the pre-trained model to avoid excessive damage to model utility.

\subsection{Mutual Information for Quantifying Concept Semantics}
Through Bayes' rule, we have $p(y|\vx)=p(\vx|y)p(y)/p(\vx)$, where $p(y)$ is fixed as the probability of erasing concept $y$. Therefore, the goal of MU is to fine-tune the pre-trained model so that $p(\vx|y)/p(\vx)$ should be as small as possible for any generated image $\vx$ by the unlearned model.  

Notice that $p(\vx|y)/p(\vx)$ is the exponent
of the mutual information between textual concept $y$ and image $\vx$, \textit{i.e.}, $\mathcal{I}(\vx,y)=\mathbb{E}_{p(\vx,y)}[\log p(\vx|y)-\log p(y)]$. It is well known that the diffusion model learns data distribution by maximizing the ELBO bound (\eqref{eq:ELBO}), which is a lower bound of sample density $p(\vx)$. Interestingly, \citet{kong2023information} establishes a rigorous connection between the pre-trained diffusion models $\theta_P$ and \textit{exact} density estimations with the help of information theory. They elucidate that the pre-trained diffusion model can well estimate the density $p(\vx)$ and $p(\vx|y)$ with the integral of optimal noise reconstruction error at different noise levels $\snr$: 
{\small\begin{align}
\displaystyle
-\log p(\vx) &=  \half \int_{0}^{\infty} \mathbb{E}_{\eps}\left[\left\|\eps-\hat{\epsilon}_{\theta_P}\left(\vx_\alpha\right)\right\|_2^2\right] d \alpha+\text{const}, \label{eq:uncond_density}\\
-\log p(\vx|y) &=  \half \int_{0}^{\infty} \mathbb{E}_{\eps}\left[\left\|\eps-\hat{\epsilon}_{\theta_P}\left(\vx_\alpha|y\right)\right\|_2^2\right] d \alpha+\text{const}. \label{eq:cond_density}
\end{align}}The derivations are introduced in Appendix~\ref{sec:inform_theoretic} for completeness and comprehension. We summarize two crucial insights to understand this relationship here: (1) The information quantity in a Gaussian noise channel is exactly the Minimum Mean Square Error (MMSE) of optimal noise reconstruction, \textit{i.e.}, $\mathcal{I}(\vx_\alpha, \vx) = \left\|\eps-\hat{\epsilon}_{\theta_P}(\vx_\alpha)\right\|_2^2$. (2) The density estimation $p(\vx)$ contains a Gaussian density term (\textit{i.e.}, the \textit{constant} term) and a \textit{correction} term (\textit{i.e.}, the \textit{first} term) that measures how much better we can denoise the target distribution than we could with optimal denoiser for Gaussian source data $p_G(\vx)\sim\mathcal{N}(0,\mathbb{I})$. Such an information-theoretic perspective demonstrates that a pre-trained model's ability to successfully reconstruct the noise from a noisy channel reflects its acquired semantic information in the images. Thus the pre-trained diffusion model is not only a powerful denoiser but also a valuable repository of semantic information. From this perspective, diminishing the generative capacity of \textit{semantic information} is equivalent to degrading their \textit{noise reconstruction} ability.



\noindent\textbf{Non-negative mutual information.} By substituting \eqref{eq:uncond_density} and \eqref{eq:cond_density} and applying orthogonality principle \cite{kay1993fundamentals}, we have a non-negative formulation of mutual information as follows:
\vspace*{-1mm}
{\small{
\begin{equation}
    \mathcal{I}(\vx,y) = \left[ \log p(\vx | y)-\log p(\vx) \right] = \half \int_{0}^{\infty} \mathbb{E}_{\eps}\left[\left\|\hat{\epsilon}_\alpha\left(\vx_\alpha\right)-\hat{\epsilon}_\alpha\left(\vx_\alpha | y\right)\right\|_2^2\right] d \alpha.
\end{equation}
}}The derivation is referred to Appendix~\ref{sec:non_neg_mi}. This non-negative expression of the mutual information has been indicated to be effective in attributing the semantics in the generated image to corresponding textual words in prompts~\cite{kong2023interpretable, dewan2024diffusion}, locating the emergence of specific semantics during generation.

\subsection{Diminishing the Mutual Information to Unlearn} 
The above information-theoretic view of diffusion model highlights that the pre-trained diffusion model is a continuous density model with competitive log-likelihood estimation, facilitating discriminating the quantity of erasing concept in generated images. Consequently, the images generated by the unlearned model should exhibit minimal mutual information with the erasing concept when inspected by the pre-trained model. This procedure is analogous to training a generative adversarial network (GAN)~\cite{goodfellow2020generative}, but the discriminator here is fixed as the pre-trained diffusion model. The gradient flow in our framework is as follows (timestep $t$ is used to indicate noise level $\alpha$ hereafter):
\begin{equation}  
\tilde \vx_t, \hat\epsilon_{\theta_U} \xrightarrow[]{\text{Generate}} \vx \xrightarrow[]{\text{Add Gaussian Noise}} \vx_t \xrightarrow[]{\text{Noise Reconstruction Prediction}} \hat\epsilon_{\theta_P}(\vx_t),\hat\epsilon_{\theta_P}(\vx_t|y) \xrightarrow[]{\text{Compute}} \mathcal{I}(\vx,y).
\end{equation}The diminish of the mutual information requires back-propagating through two forward paths of the generator $\theta_U$ and discriminator $\theta_P$, which is computationally expensive and impractical for large foundation models. In this section, we analyze the gradient flow of the above optimization and propose an alternative objective.

We denote the mutual information at timestep $t$ as $\mathcal{I}_t(\vx,y):=\half\left\|\hat{\epsilon}_{\theta_P}\left(\vx_t|y\right)-\hat{\epsilon}_{\theta_P}\left(\vx_t\right)\right\|_2^2$, and then minimizing the unlearning objective in \eqref{eq:forget} corresponds to minimize each $\mathcal{I}_t(\vx,y)$. The gradient of $\mathcal{I}_t(\vx,y)$ w.r.t. the unlearned model $\theta_U$ is as the following: 
{\small{\begin{equation}
\begin{aligned}
     \dfrac{\partial\mathcal{I}_t(\vx,y)}{\partial \theta_U} &= \dfrac{ \partial \mathcal{I}_t(\vx,y)}{\partial \hat \epsilon_{\theta_P}}\cdot\dfrac{\partial \hat \epsilon_{\theta_P}}{\partial \vx_t} \cdot \dfrac{\partial \vx_t}{\partial \vx} \cdot \dfrac{\partial \vx}{\partial \theta_U}\\      &=\mathbb{E}_{\eps}\biggl[w(t)\cdot \underbrace{(\hat{\epsilon}_{\theta_P}\left(\vx_t|y\right)-\hat{\epsilon}_{\theta_P}\left(\vx_t\right))}_{\text{Pre-trained CFG}}\cdot\underbrace{(\dfrac{\partial \hat{\epsilon}_{\theta_P}\left(\vx_t|y\right) }{\partial \vx_t}-\dfrac{\partial \hat{\epsilon}_{\theta_P}\left(\vx_t\right)}{\partial \vx_t})}_{\text{Pre-trained U-Net Jacobian}}\cdot\underbrace{\dfrac{\partial \hat\epsilon_{\theta_U}(\tilde \vx_t|y)}{\partial \theta_U}}_{\text{Unlearned U-Net Gradient }}\biggr].
\end{aligned}
\label{eq:mutual_t_grad}
\end{equation}}}The derivation is referred to Appendix~\ref{sec:theory_mi_grad} and $w(t)$ is a coefficient. The obtained gradient in \eqref{eq:mutual_t_grad} can be decomposed into 3 components: (1) CFG term of the pre-trained diffusion model, (2) U-Net Jacobian of the pre-trained diffusion model and (3) U-Net gradient of the unlearned model.

\textbf{Kullback-Leibler (KL) divergence minimization by omitting the U-Net Jacobian.} 
The Jacobian term in \eqref{eq:mutual_t_grad} is computationally expensive to evaluate and poorly conditioned at low noise levels, as it approximates the scaled Hessian of the marginal density. Following common practice in Score Distillation Sampling (SDS) \cite{poole2022dreamfusion, wang2023score}, we omit this term and obtain an approximate gradient:
{\small{\begin{equation}
\label{eq:kl_distill_grad}
\frac{\partial \mathcal{I}_t(\vx, y)}{\partial \theta_U} 
\approx \mathbb{E}_\eps\biggl[w(t)\cdot\underbrace{(\hat\epsilon_{\theta_P}(\vx_t|y) - \hat\epsilon_{\theta_P}(\vx_t))}_{ \text{Pre-trained CFG}}\cdot\underbrace{\dfrac{\partial \hat\epsilon_{\theta_U}(\tilde \vx_t|y)}{\partial \theta_U}}_{\text{Unlearned U-Net Gradient}} \biggr].
\end{equation}}}Then we show that the approximate gradient in \eqref{eq:kl_distill_grad} corresponds to minimizing the KL divergence between the unconditional and conditional latent distributions at timestep \( t \):
\vspace{-1mm}
{\small{\begin{equation}
\label{eq:kl_distill}
\minimize_{\theta_U}\;\mathcal{D}_{\theta_P}^{\mathrm{KL}}(\vx) := \mathrm{KL}(p_{\theta_P}(\vx_t) \| p_{\theta_P}(\vx_t|y)) = \mathbb{E}_{\eps} \left[ \log \dfrac{p_{\theta_P}(\vx_t)}{p_{\theta_P}(\vx_t|y)}\right].
\end{equation}}}The derivation is referred to Appendix~\ref{sec:theory_kl_equivalence}. Intuitively, the unlearning objective in \eqref{eq:forget} is converted to demanding the unlearned model to shift its generation \textit{at any noise level} towards a density region less associated with the concept $y$, where relevance is inspected by the pre-trained diffusion model.


\subsection{Deviating Least from the Pre-trained Model to Preserve}\label{sec:least_deviation}
Directly minimizing the KL divergence in \eqref{eq:kl_distill} w.r.t. the model parameters $\theta_U$ overlooks the preservation of innocent generations, generally resulting in a severe performance degradation. To address this, we seek the minimums of \eqref{eq:kl_distill} for the one with favorable model utility. The minimum is attained when the expected log-ratio $\log p_{\theta_P}(\vx_t|y)/\log p_{\theta_P}(\vx_t)=0$.
The optimal sampling distribution of the unlearned model is the one that removes concept-related generations meanwhile maintaining other generations, thus, we prioritize degrading the conditional distribution $p_{\theta_U}(\vx|y)$. The collection of concept-independent conditional distribution for \eqref{eq:kl_distill} is defined as:
{\small{\begin{equation}
\label{eq:optimal_sampling}
\mathcal{Q}^*(t,y) := \left\{q_{\theta_U}(\vx|y) \,\middle|\, p_{\theta_P}(\vx_t|y) = p_{\theta_P}(\vx_t),\ \forall \vx \sim q_{\theta_U}(\vx|y)\right\}.
\end{equation}}}When erasing the undesired concept, we should best preserve the generation quality over the rest of the prompt space. \textit{Therefore, we seek a $q_{\theta_U}(\vx|y)\in Q^*$ that is closest to $p_{\theta_P}(\vx|y)$ to avoid excessive forgetting}. In information theory, among all the distributions that are independent of $y$, the one that has minimal KL divergence to $p_{\theta_P}(\vx_t|y)$ is the marginal distribution $p_{\theta_P}(\vx_t)$ \cite{amari2016information}:
{\small{\begin{equation}
    \minimize_{q_{\theta_U}(\vx_t|y)\in Q^*}\quad \mathbb{E}_{\vx\sim p_{\theta_U}}\ [KL(q_{\theta_U}(\vx_t|y)\|p_{\theta_P}(\vx_t|y))] 
    \ \Longrightarrow 
    q_{\theta_U}^*(\vx_t|y)= p_{\theta_P}(\vx_t).
\end{equation}}}In this way, we can have minimal impact on other generations when effectively erasing the concept. Therefore, we fine-tune the unlearned model to approach its conditional generation $p_{\theta_U}(\vx|y)$ to approach the marginal distribution of the pre-trained diffusion model $p_{\theta_P}(\vx)$. Mathematically, aligning these two distributions is equivalent to aligning the conditional score of the unlearned model with the unconditional one of the pre-trained model:
{\small{
\begin{equation}
   \minimize_{\theta_U}\ {\mathcal{D}_{\text{MI}}}(\vx):= \mathrm{KL}(q^*_{\theta_U}(\vx_t|y)\|q_{\theta_U}(\vx_t|y)) = \mathbb{E}_{\eps} \left[\left\| \hat\epsilon_{\theta_U}(\vx_t|y) -\hat\epsilon_{\theta_P}(\vx_t)\right\|_2^2\right]
\label{eq:MUMI_loss}
\end{equation}}}The derivation is referred to Appendix~\ref{sec:theory_mim_mu}. In practice, we use a fixed set of concept-related images $\mathcal{X}_f$ to stand for the conditional sampling distribution of unlearned model $q_{\theta_U}(\vx|y)$, \textit{i.e.}, $\vx_t = \sqrt{\bar{\alpha}_t} \vx+\sqrt{1-\bar{\alpha}_t} \eps,\,\vx\in\mathcal{X}_f$. \textit{From the information-theoretic perspective, such alignment degrades the conditional denoiser's ability to reconstruct semantic information from noisy channels, diminishing the acquired information quantity in model parameters.}  


\textbf{Re-interpret Safe-Self-Distillation(SDD) \cite{kim2023towards}.} 
At the first sight, our proposed \eqref{eq:MUMI_loss} is quite similar to SDD~\cite{kim2023towards} in formulation. SDD aligns the unlearned model's conditional score with its \textit{own} unconditional one:
\vspace*{-1mm}
{\small\begin{equation}
     \displaystyle \minimize_{\theta_U} ~ \mathcal{D}_{\text{SDD}}(\vx):=\mathbb{E}_{t} \left[\left\| \hat\epsilon_{\theta_U}(\tilde \vx_t|y) -\hat\epsilon_{\theta_U}(\tilde \vx_t)\right\|_2^2\right], \quad \tilde \vx_t = \vx_T + \sum_{k=T}^{t+1} \hat\epsilon_{\theta_{\text{EMA}}}(\tilde \vx_k|y).
\label{eq:sdd}
\end{equation}}
\vspace*{-3mm}

\noindent where $\vx_T\sim\mathcal{N}(0,\mathbb{I})$ and $\theta_{\text{EMA}}$ is an exponential moving average (EMA) updated version of the unlearned model. Regrettably, this intuitively reasonable alignment did not receive much attention. 
Now by comparing SDD with \eqref{eq:kl_distill_grad} and our \eqref{eq:MUMI_loss}, we found the 
reasons that might lead to sub-optimal performance. 
In SDD,  the gradient of $\mathcal{D}_{\text{SDD}}$ w.r.t. the unlearned model is:
\vspace*{-1mm}
{\small{\begin{equation}
    \dfrac{\partial \mathcal{D}_{\text{SDD}}(\vx)}{\partial \theta_U}=\mathbb{E}_\eps\biggl[ \underbrace{(\hat\epsilon_{\theta_U}(\tilde \vx_t|y) -  \hat\epsilon_{\theta_U}(\tilde \vx_t))}_{\text{Unlearned CFG}}
    \dfrac{\partial \hat\epsilon_{\theta_U}(\tilde \vx_t|y)}{\partial \theta_U}\biggr].
    \label{eq:sdd_grad}
\end{equation}}}Here, the CFG term 
is provided by the unlearned model, whose discriminability of the erasing concept might gradually become deficient with the gone of generation capability regarding the erasing concept. Besides, to preserve model utility on other generations, the unlearned distribution should deviate least from the original one when erasing. While the distillation teacher in SDD is $p_{\theta_U}(\vx)$, which drifts further and further away from $p_{\theta_P}(\vx|y)$ as unlearning proceeds. In Sec.~\ref{sec:compensation_free_comparison}, we empirically verify that the model utility of the SDD unlearned model gradually breaks down when ceaselessly
self-distilling from its own unconditional distribution.

\section{Experiments}
\subsection{Experiment Setups}
\label{sec:exp_setup}

\noindent\textbf{Evaluation models and datasets.} There are mainly two kinds of concept erasing: (1) forget an artistic painting \textit{style} for the purpose of copyright protection; (2) forget an \textit{object} for the degeneration of models. We evaluate on a comprehensive benchmark, \textbf{UnlearnCanvas}~\cite{zhang2024unlearncanvas}. It includes 70 concepts in total, which are divided into two main domains: a style domain comprising 50 artist styles and an object domain with 20 common objects. Each image in the dataset is a stylized object, \textit{e.g.}, ``A Dog in Van Gogh Style''. The variety of concepts enables a comprehensive evaluation over a rich unlearning target bank, and the stylized object allows a distinct inspection on the retainability of innocent knowledge from both in-domain and cross-domain perspectives. The pre-trained model in UnlearnCanvas benchmark is stable-diffusion-v1-5 fine-tuned on UnlearnCanvas dataset.

\textbf{Evaluation metrics.}
Machine unlearning in generative models should be assessed from three perspectives: the \textit{completeness} of the erasure, the \textit{retainability} of innocent knowledge, and the \textit{quality} of generated images. We use Unlearning Accuracy (\textbf{UA}), \textit{i.e.}, the proportion of generated images classified as the erasing concept, to indicate unlearning completeness. For retainability, we indicate with In-domain Retain Accuracy (\textbf{IRA}) and Cross-domain Retain Accuracy (\textbf{CRA}). For example, when unlearning ``Van Gogh'' style, IRA refers to successful generations of other styles (\textit{e.g.}, ``Monet'') and CRA refers to successful generations of all the objects (\textit{e.g.}, ``Dogs''). We indicate the distributional quality of image generation with Fréchet Inception Distance (\textbf{FID}). Furthermore, most existing erasing methods rely on explicitly compensating for a presupposed scope of remaining concepts, where the retainability is typically evaluated. However, we argue that retainability should also be assessed in an out-of-distribution (O.O.D) concept domain--\textit{i.e.}, concepts unseen during unlearning--to ensure the \textit{general} utility of the unlearned model. 

\noindent\textbf{Unlearning baselines.} 
Original UnlearnCanvas benchmark contains 9 most recently proposed MU methods for diffusion models, including (1) \textbf{ESD} (Erased Stable Diffusion) \cite{gandikota2023erasing}, (2) \textbf{FMN} (Forget-Me-Not) \cite{zhang2024forget}, 
(3) \textbf{CA} (Ablating Concepts) \cite{kumari2023ablating}, (4) \textbf{UCE} (Unified Concept Editing) \cite{gandikota2024unified}, (5) \textbf{SalUn} (saliency Unlearning) \cite{fan2023salun}, (6) \textbf{SEOT} (Suppress EOT) \cite{li2024get}, (7) \textbf{SHS} (ScissorHands) \cite{wu2024scissorhands}, (8) \textbf{EDiff} (EraseDiff) \cite{wu2024erasediff}, and (9) \textbf{SPM} (concept-SemiPermeable Membrane) \cite{lyu2023one}. We add (10) \textbf{SDD} (Safe Self-distilling) for further comparison due to its benign retainability. 



\subsection{Unlearning Performance on UnlearnCanvas Benchmark} 
\textbf{Tab.}~\ref{tab:unlearncanvas} demonstrates performance of style and object unlearning in UnlearnCanvas benchmark. To better distinguish the locality of existing erasing operations, we divide existing methods into two groups based on whether explicit compensations for innocent concepts are applied, with the bottom group representing \textit{compensation-free} methods. It is evident in \textbf{Tab}.~\ref{tab:unlearncanvas} that existing methods face great challenges with either incomplete unlearning (\textit{e.g.}, CA, SEOT, SPM, FMN) or severe damage to innocent knowledge (\textit{e.g.}, ESD, FMN, UCE, SHS, EDiff) as highlighted by \warn{red} color. In terms of accuracy metrics, SalUn achieves the best total average accuracy ($92.77\%$), appearing to strike a good balance between unlearning and retention. However, we reveal in our subsequent experiment that re-assimilating the remaining data in SalUn is limited, and SalUn exhibits poor retainability in concepts out of the compensation scope.

Among the 3 existing compensation-free methods, SDD achieves a total average accuracy of $81.00\%$, surpassing all the other compensation-dependent methods except for SalUn and EDiff.
While our method further improves upon SDD with a substantial accuracy gain of \textbf{$8.42\%$}. Noteworthily, our IRA and CRA consistently exceed $90\%$, indicating a favorable retainability. Moreover, our method achieves the lowest FID at $49.14$, while that of SalUn and SDD are $61.05$ and $70.40$ respectively. The significant gap between SalUn and MiM-MU in FID suggests that post-remedial compensations might have limited ability to fully restore the original generation quality. 

\begin{table*}[htbp]
\vspace*{-2mm}
\centering
\caption{Quantitative performance overview of different DM unlearning methods on UnlearnCanvas benchmark~\cite{zhang2024unlearncanvas}. Methods are divided into two categories depending on whether there is explicit maintenance for the remaining concepts.  Each unlearning request targets either a single style or a single object, and the results are averaged across 50 styles and 20 objects, respectively. Best values are marked in \congrat{green}, second-best values are marked in \secbest{yellow} for Avg.Acc and FID columns, and underperforming ones are marked in \warn{red} for each column.}
\label{tab:unlearncanvas}
\resizebox{\textwidth}{!}{%
\begin{tabular}{lcccccccccc}
\toprule
\textbf{Method} & 
\multicolumn{4}{c}{\textbf{Style Unlearning}} & 
\multicolumn{4}{c}{\textbf{Object Unlearning}} & 
\textbf{Total } & 
\textbf{FID $\downarrow$} \\
\cmidrule(lr){2-5} \cmidrule(lr){6-9}
& \textbf{UA $\uparrow$} & \textbf{IRA $\uparrow$} & \textbf{CRA $\uparrow$} & \textbf{Avg. Acc $\uparrow$} 
& \textbf{UA $\uparrow$} & \textbf{IRA $\uparrow$} & \textbf{CRA $\uparrow$} & \textbf{Avg. Acc $\uparrow$} 
& \textbf{Avg. Acc $\uparrow$} & \\
\midrule
CA \cite{kumari2023ablating}   
& \warn{${60.82\%}$} & ${96.01}\%$ & $92.70\%$ & $83.84\%$
& \warn{${46.67\%}$} & $90.11\%$ & $81.97\%$ & $72.92\%$ 
& $78.38\%$ & ${54.21}$ \\

UCE \cite{gandikota2024unified}  
& ${98.40}\%$ & \warn{${60.22\%}$} & \warn{${47.71\%}$} & $68.78\%$ 
& ${94.31}\%$ & \warn{${39.35\%}$} & \warn{${34.67\%}$} & \warn{$56.11\%$} 
&\warn{ $62.45\%$} & \warn{${182.01}$} \\

SEOT \cite{li2024get} 
& \warn{${56.90\%}$} & $94.68\%$ & $84.31\%$ & $78.63\%$ 
& \warn{${23.25\%}$} & $95.57\%$ & $82.71\%$ & $67.18\%$ 
& $72.91\%$ & $62.38$ \\

SalUn \cite{fan2023salun} 
& $86.26\%$ & $90.39\%$ & $95.08\%$ & \secbest{$90.58\%$} 
& $86.91\%$ & ${96.35}\%$ & ${99.59}\%$ & \congrat{$94.95\%$}
& \congrat{$92.77\%$} & $61.05$ \\

SPM \cite{lyu2023one} 
& \warn{${60.94\%}$} & $92.39\%$ & $84.33\%$ & $79.22\%$ 
& $71.25\%$ & $90.79\%$ & $81.65\%$ & $81.23\%$ 
& $80.23\%$ & \secbest{$59.79$} \\

EDiff \cite{wu2024erasediff} 
& $92.42\%$ & ${73.91\%}$ & ${98.93}\%$ & $88.42\%$ 
& $86.67\%$ & $94.03\%$ & \warn{${48.48\%}$} & $76.39\%$ 
& $82.41\%$ & $81.42$ \\

SHS \cite{wu2024scissorhands} 
& $95.84\%$ & $80.42\%$ & \warn{${43.27\%}$} & $73.18\%$ 
& $80.73\%$ & $81.15\%$ & $67.99\%$ & $76.63\%$ 
& $74.91\%$ & \warn{119.34} \\

\midrule

ESD \cite{gandikota2023erasing} 
& ${98.58}\%$ & $80.97\%$ & $93.96\%$ &  \congrat{$91.17\%$} 
& $92.15\%$ & \warn{${55.78\%}$} & \warn{${44.23\%}$} & $64.05\%$ 
& $77.61\%$ & $65.55$ \\

FMN \cite{zhang2024forget}  
& $88.48\%$ & \warn{${56.77\%}$} & \warn{${46.60\%}$} & \warn{$63.95\%$} 
& \warn{${45.64\%}$} & $90.63\%$ & $73.46\%$ & $69.91\%$ 
& $66.93\%$ & \warn{${131.37}$} \\

SDD \cite{kim2023towards} 
& $83.79\%$ & $75.28\%$ & $77.38\%$ & $78.82\%$ 
& $84.31\%$ & $79.71\%$ & $85.51\%$ & $83.18\%$ 
& $81.00\%$ & $70.40$ \\

\rowcolor{Gray}
MiM-MU & $80.12\%$ & $93.99\%$ & $93.18\%$ & $89.10\%$ & $81.14\%$ & $91.41\%$ & $96.65\%$ & \secbest{$89.73\%$} & \secbest{$89.42\%$} & \congrat{49.14} \\
\bottomrule[1pt]
\end{tabular}}
\end{table*}

\begin{figure}[htbp]
  \centering
  \includegraphics[width=0.98\linewidth]
  {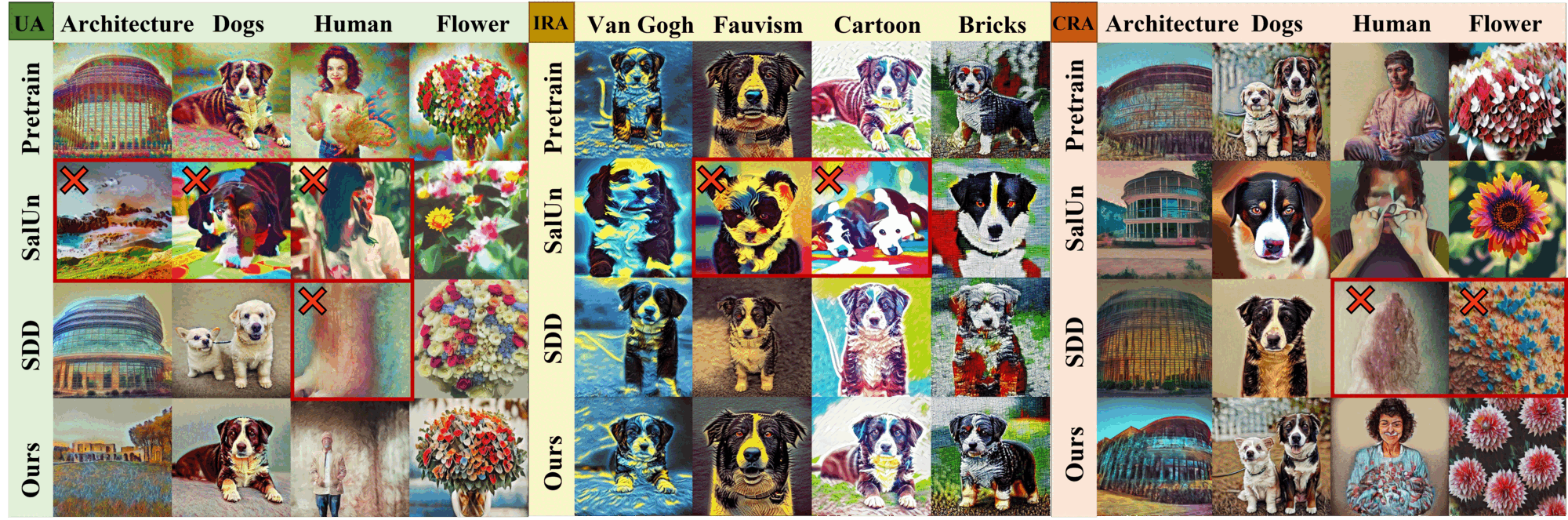}
    \caption{
    \footnotesize \textbf{Qualitative performance overview of different DM unlearning methods on ``Monet'' style.} For unlearning completeness (\textbf{UA}), we showcase the erasing concept with 4 concepts from the cross domain, \textit{i.e.}, 4 different objects in ``Monet'' style. For the retainability, we use ``Dogs'' object 
 and ``Seed Image'' style to combine with 4 remaining in-domain (\textbf{IRA}) and cross-domain concepts (\textbf{CRA}). Generations of the unlearned model should be \textit{different} from the pre-trained model in terms of the erasing concept in \textbf{UA} and should \textit{approach} the pre-trained model in \textbf{IRA} and \textbf{CRA}.} 
  \label{fig:UnlearnCanvas}
\vspace*{-6mm}
  
\end{figure}

The accuracy-related metrics reported above primarily indicate the presence or absence of a concept, while a more detailed evaluation of concept generation quality requires examining the generated images. \textbf{Fig.~\ref{fig:UnlearnCanvas}} showcases generations of ``Monet'' style unlearned methods (More visualizations are referred to Appendix~\ref{sec:unlearncanvas_vis}). It can be observed that SalUn fails to generate certain objects when requested alongside the ``Monet'' style in UA, indicating its excessive degradation. Moreover, its generations for the remaining concepts exhibit noticeable distortions and color oversaturation. Both implies that Salun's damage on image generation lacks of specificity and re-assimilating the remaining data have limited effectiveness in restoring fine-grained details. In contrast, MiM-MU faithfully generates the object without any painting style in UA and produces high-quality images (\textit{i.e.}, nearly identical to those generated by the pre-trained model) in IRA and CRA, demonstrating a more nuanced removal. \textbf{We think this might be an encouraging progress which demonstrates that refraining from invasive interference during erasure is a more effective and reliable strategy for preserving the generation quality of large-scale generative models than post-remedial compensations.}  


\subsection{Compensation-free vs. Compensation-dependent MU Methods}\label{sec:compensation_free_comparison}
While SalUn achieves superior performance on accuracy-related metrics in the UnlearnCanvas benchmark, its generation quality (FID) is notably inferior to ours. This suggests that its compensation mechanisms fail to fully recover the original generation quality. Furthermore, we highlight two fundamental \textbf{limitations} that critically undermine the reliability of post-remedial compensations: (1) The inadvertent damage caused by indiscriminate removal is difficult to diagnose, potentially creating subtle but \textbf{cumulative performance degradation that remains unrepaired}. (2) While generative models must handle a vast and diverse range of concepts, current compensation is typically restricted to a presupposed scope, which is relatively minimal so that \textbf{generations beyond the scope might remain poor}. We empirically validate these limitations through: (1) sequential unlearning tasks and (2) retainability assessment on out-of-distribution (O.O.D.) concept domain.

\textbf{Sequential unlearning (SU).} In practice, unlearning requests might arrive sequentially, demanding multiple executions of unlearning methods. Extensive evaluations in UnlearnCanvas~\cite{zhang2024unlearncanvas} benchmark (Tab. A7 in original paper) demonstrate that \textbf{none of} existing MU methods could achieve resilient unlearning performance during SU. There are 3 universal deficiencies: \textbf{(i)} \textit{degraded retainability}, \textbf{(ii)} \textit{unlearning rebound effect}, and \textbf{(iii)} \textit{catastrophic retaining failure}. These phenomena imply that the undesired knowledge is not essentially removed and post-remedial compensation is disordered, exposing fragile knowledge management (\textit{i.e.}, removal and retention) of existing compensation-dependent methods. We sequentially unlearn 6 artist styles (in line with UnlearnCanvas) and track unlearning and retaining performances for each unlearning request in \textbf{Tab.}~\ref{tab:sequential_mu} (while visualizations are provided in Appendix~\ref{sec:sequential_unlearn}). We highlight with \warn{red} that the unlearned model's performance on previous erased concept stages a recovery in SalUn, which implies that the erased concept is merely temporarily concealed. In contrast, our UA of each unlearning request stays at a high level throughout SU, indicating superior resilience of MiM-MU. Moreover, without any compensation, our RA is consistently higher than SalUn during SU. The resilient unlearning and benign retention collectively demonstrate the promising knowledge management of the compensation-free method, MiM-MU.

\begin{table*}[htbp]
\centering
\caption{\footnotesize Quantitative performance of MiM-MU and SalUn in sequential unlearning (SU) task. Each column represents a new unlearning request, denoted by $\mathcal{T}_i$, where $\mathcal{T}_1$ is the oldest. Each row represents the UA for a specific unlearning objective or the retaining accuracy (RA), given by the average of IRA and CRA. Results indicating \textit{unlearning rebound} effect (where $\Delta \text{UA}\geq 5\%$) are highlighted in  \warn{red}.}
\setlength\tabcolsep{4pt}
\footnotesize
\label{tab:sequential_mu}
\resizebox{\linewidth}{!}{
\begin{tabular}{lccccccclccccccc}
\toprule
\multicolumn{8}{c}{\textbf{MiM-MU}} & \multicolumn{8}{c}
{\textbf{SalUn}} \\
\midrule 
\multicolumn{2}{l}{\multirow{2}{*}{\textbf{Metrics}}} 
& $\mathcal{T}_1$ & $\mathcal{T}_1\sim\mathcal{T}_2$ & $\mathcal{T}_1\sim\mathcal{T}_3$ & $\mathcal{T}_1\sim\mathcal{T}_4$ & $\mathcal{T}_1\sim\mathcal{T}_5$ & $\mathcal{T}_1\sim\mathcal{T}_6$ 
& \multicolumn{2}{l}{\multirow{2}{*}{\textbf{Metrics}}} 
& $\mathcal{T}_1$ & $\mathcal{T}_1\sim\mathcal{T}_2$ & $\mathcal{T}_1\sim\mathcal{T}_3$ & $\mathcal{T}_1\sim\mathcal{T}_4$ & $\mathcal{T}_1\sim\mathcal{T}_5$ & $\mathcal{T}_1\sim\mathcal{T}_6$ \\ 
& & \multicolumn{6}{c}{-------------------------- \textbf{Unlearning Request} --------------------------}
& & & \multicolumn{6}{c}{-------------------------- \textbf{Unlearning Request} --------------------------} \\
\cmidrule(r){1-8} \cmidrule(l){9-16}
\multirow{6}{*}{\textbf{UA $\uparrow$ }} & $\mathcal{T}_1$  & 98.00\% & 99.00\% & 98.00\% & 99.00\% & 100.00\% & 100.00\% 
& \multirow{6}{*}{\textbf{UA $\uparrow$ }} & $\mathcal{T}_1$  & \warncellred{84.00\%} & \warncellred{79.00\%} & \warncellred{78.00\%} & \warncellred{65.00\%} & 67.00\% & 64.00\% \\
& $\mathcal{T}_2$ & - & 93.00\% & 96.00\% & 96.00\% & 97.00\% & 100.00\% 
& &$\mathcal{T}_2$ & - & \warncellred{81.42\%} & \warncellred{75.00\%} & 72.00\% & \warncellred{69.00\%} & \warncellred{61.00\%} \\
&$\mathcal{T}_3$ &  - & - & 97.00\% & 97.00\% & 100.00\% & 99.00\% 
& &$\mathcal{T}_3$ & - & - & \warncellred{90.00\%} & \warncellred{85.00\%} & 84.00\% & 87.00\% \\
&$\mathcal{T}_4$ &  - & - & - & 98.00\% & 100.00\% & 100.00\% 
& &$\mathcal{T}_4$ & - & - & - & 84.00\% & \warncellred{86.00\%} & \warncellred{81.00\%} \\
&$\mathcal{T}_5$ &  - & - & - & - & 100.00\% & 100.00\% 
& & $\mathcal{T}_5$& - & - & - & - & 79.00\% & 81.00\% \\
&$\mathcal{T}_6$ & - & - & - & - & - & 92.00\% 
& & $\mathcal{T}_6$ & - & - & - & - & - & 89.00\% \\
\midrule
\textbf{RA $\uparrow$} & & 90.90\% & 86.56\% & 80.58\% & 77.27\% & 70.34\% & 67.47\% 
& & & 85.43\% & 80.32\% & 71.42\% & 65.41\% & 63.24\% & 60.19\% \\
\bottomrule[1pt]
\end{tabular}}
\end{table*}

\begin{figure}[htbp]
  \centering
  \includegraphics[width=0.9\linewidth]
  {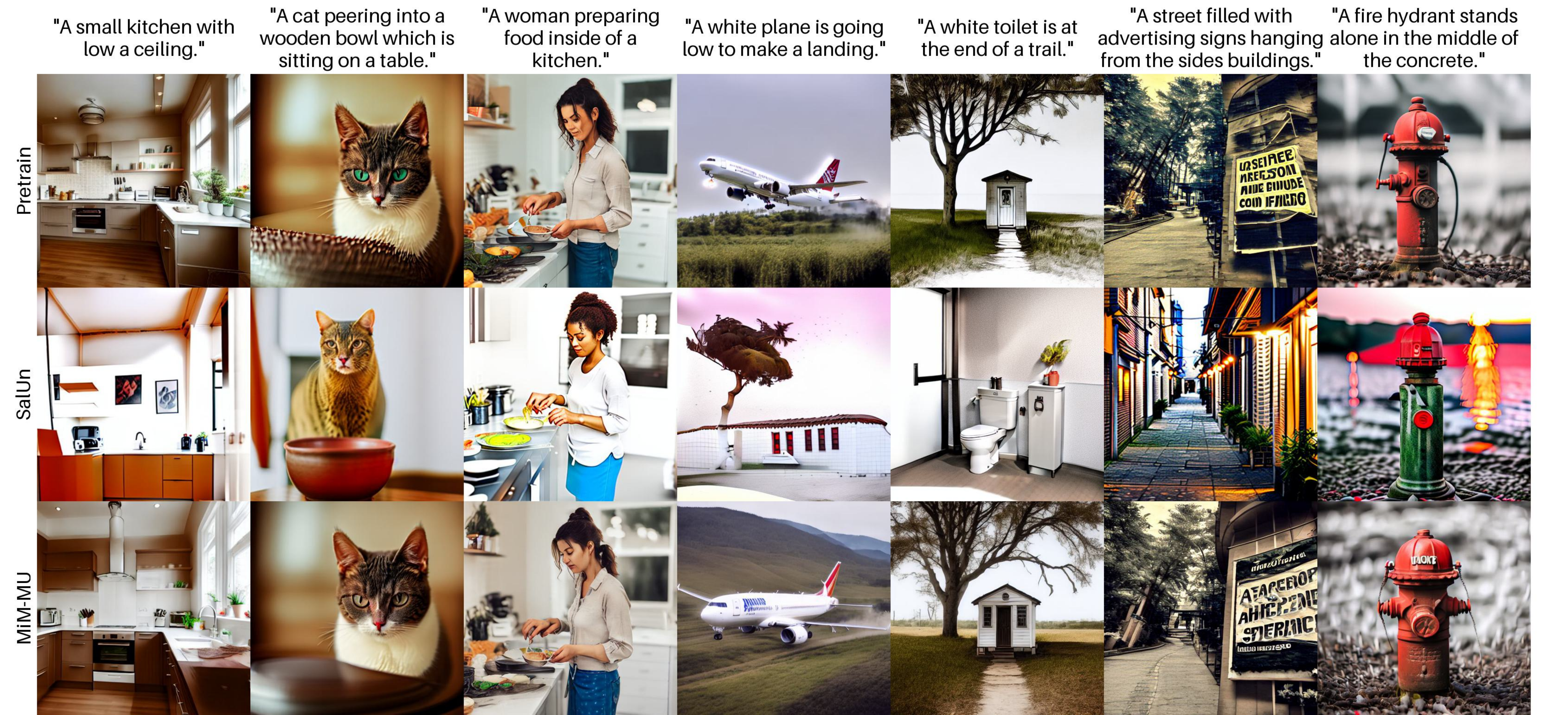}
  \caption{
    \footnotesize \textbf{Qualitative visualizations of MiM-MU and SalUn unlearned model on COCO-10k dataset generations.} Generations of SalUn unlearned model exhibit distortions and misalignment with the requested prompts. In contrast, MiM-MU produces high-quality and textually aligned images on COCO-10k dataset without utilizing additional maintenance to preserve model utility.}
  \label{fig:ood_coco}
\vspace*{-6mm}
\end{figure}


\begin{wraptable}[6]{r}{0.28\textwidth}
\centering
\vspace*{-5mm}
\caption{
\footnotesize Generation quality in COCO-10k datasets. }
\vspace{-1mm}
\setlength\tabcolsep{4.0pt} 
\footnotesize
\begin{tabular}{lcc}
\toprule
\textbf{Method} & \textbf{FID $\downarrow$} & \textbf{CLIP $\uparrow$} \\
\midrule
Pretrain & 37.31 & 0.2780 \\
SalUn    & 49.74 & 0.2761 \\
Ours     & \textbf{34.42} & \textbf{0.2780} \\
\bottomrule
\end{tabular}
\label{tab:coco_domain}
\end{wraptable}
\textbf{Deteriorated general utility in O.O.D. Domain Concepts of the compensation-dependent method.} We empirically validate our concern about O.O.D. retainability by comparing MiM-MU with SalUn on COCO-10k dataset. Quantitative metrics are reported in \textbf{Tab.}~\ref{tab:coco_domain} and qualitative visualizations are showcased in \textbf{Fig.}~\ref{fig:ood_coco}. The ``Pretrain'' in \textbf{Tab.}~\ref{tab:coco_domain} refers to the stable-diffusion-v1-5 fine-tuned on UnlearnCanvas dataset, which has worse FID than vanilla sd-v1.5 due to learning stylized images. Notably, SalUn yields a significantly higher FID than the pre-trained model, indicating its poor retainability in COCO-10k generations. In contrast, MiM-MU achieves an even lower FID than the pre-trained model, suggesting that its erasure not only preserves utility but also successfully recovers the quality degradation introduced by fine-tuning. Visualizations in \textbf{Fig.}~\ref{fig:ood_coco} demonstrate that MiM-MU generates more textually aligned images with clear textures, whereas generations of SalUn suffer from text-image misalignment and pronounced distortions. This result underscores the advantage of a nuanced erasure to ensure the \textit{general} utility after unlearning.

\textbf{Compensation-free method exhibits advantage in fine-grained erasure.}
Although UnlearnCanvas benchmark represents a significant step forward compared to earlier datasets, it still adopts relatively coarse-grained concept definitions, where concepts are relatively independent (\textit{e.g.}, Dogs and Cats). However, in real-world settings, semantic boundaries are often highly entangled. To further assessing the granularity of existing concept erasure methods, we perform fine-grained erasure over 3 fine-grained datasets, Standford Dogs~\cite{Stanford_Dogs}, Oxford Flowers~\cite{Nilsback08} and CUB-200~\cite{welinder2010caltech}. To emphasize the potential risk of existing careless removal, we only compensate (\textit{i.e.}, replaying corresponding data) generations of neighborhood classes in SalUn, while leaving remaining classes (termed \textit{Other Concepts}) uncompensated, revealing their unintended and unrevealed degradation to other concepts.

\begin{table*}[htbp]
\vspace*{-2mm}
\centering
\caption{\footnotesize Quantitative comparison between SalUn and Mim-MU on 3 fine-grained classification datasets, Standford Dogs~\cite{Stanford_Dogs}, Oxford Flowers~\cite{Nilsback08} and CUB-200~\cite{welinder2010caltech}. Best values are marked in \congrat{green} and under-performing ones are marked in \warn{red} for each column.}
\label{tab:finegrained_erasure}
\resizebox{0.88\textwidth}{!}{%
\begin{tabular}{lcccccccccc}
\toprule
\textbf{Dataset} & \textbf{Method} & \textbf{UA $\uparrow$} & \textbf{NRA $\uparrow$} & \textbf{ORA $\uparrow$} & \textbf{Avg.Acc $\uparrow$} & \textbf{N.FID $\downarrow$} & \textbf{O.FID $\downarrow$} \\
\midrule
\multirow{2}{*}{Dogs} 
& SalUn   & $95.83\%$ & $35.83\%$ & $65.61\%$ & $65.76\%$ & $108.19$ & $54.06$ \\
& Mim-MU  & $100.00\%$ & $37.50\%$ & $70.00\%$ & \congrat{$69.17\%$} & \congrat{$92.46$}  & \congrat{$42.42$} \\
\midrule
\multirow{2}{*}{Flowers} 
& SalUn   & \warn{$66.67\%$} & $44.17\%$ & $67.81\%$ & $59.55\%$ & $131.83$ & $64.93$ \\
& Mim-MU  & $100.00\%$ & $43.33\%$ & $68.54\%$ & \congrat{$70.62\%$} & \congrat{$130.99$} & \congrat{$58.51$} \\
\midrule
\multirow{2}{*}{CUB-200} 
& SalUn   & $91.67\%$ & $78.33\%$ & $68.30\%$ & \congrat{$79.43\%$} & $52.56$  & \warn{$42.04$} \\
& Mim-MU  & $95.83\%$ & \warn{$47.50\%$} & $67.42\%$ & $70.25\%$ & \congrat{$56.61$}  & \congrat{$16.30$} \\
\bottomrule[1pt]
\vspace{-3mm}
\end{tabular}}
\end{table*}

Performances are referred to \textbf{Tab.}\ref{tab:finegrained_erasure}. Across all 3 datasets, Mim-MU consistently achieves more thorough forgetting  and SalUn fails in Oxford Flower dataset, with UA only $66.67\%$ while Mim-MU is $100.0\%$.  The retainability include 2 perspectives in fine-grained erasure: Neighborhood Retain Accuracy (NRA) and Other-class Retain Accuracy (ORA). For near classes, Mim-MU fails to preserve generation of neighboorhood classes in CUB-200, with an NRA of $47.50\%$ while that of SalUn is $78.33\%$. In the rest 2 datasets, Mim-MU achieves comparable retainability with accuracy floating within $\pm 2\%$. It is worth noting that we explicitly compensate the neighborhood classes for SalUn, therefore, we regard it credible for Mim-MU to achieve comparable retainability. For ORA, Mim-MU achieves better retainability in these, which are not explicitly compensated in SalUn as well. Consistently,  Mim-MU exhibits lower FID scores in both neighborhood and other concepts across 3 datasets, indicating its stronger capability to maintain high-quality image generation and prevent undesired artifacts.

\begin{figure}[htbp]
  \centering
\includegraphics[width=0.98\linewidth]
  {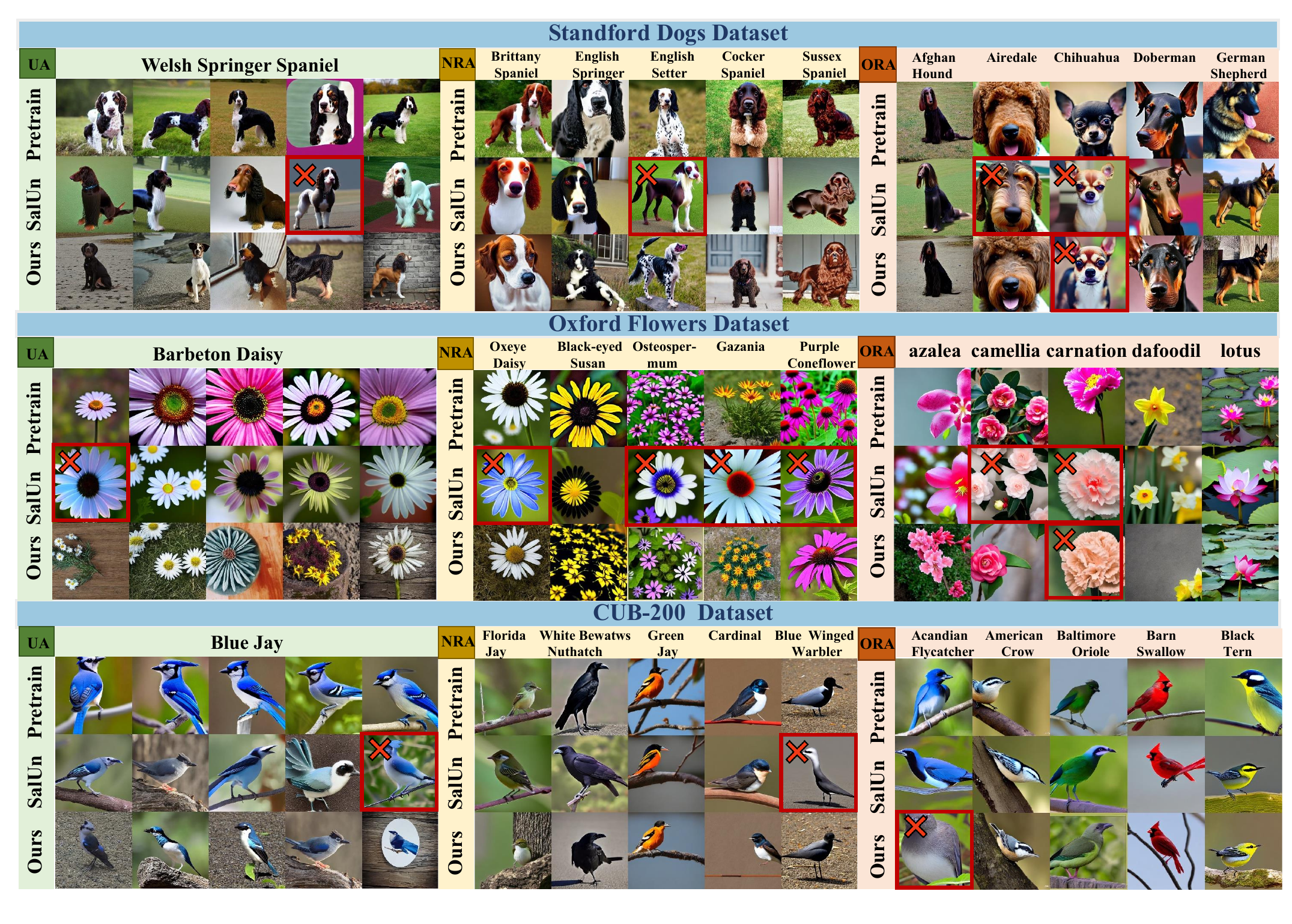}
    \caption{
    \footnotesize \textbf{Qualitative performance overview of Mim-MU and SalUn on 3 fine-grained datasets.} For unlearning completeness (\textbf{UA}), we showcase the erasing concept with 5 seeds. For the retainability, we indicate 5 neighborhood concepts (\textbf{NRA}) and 5 other concepts (\textbf{ORA}). Generations of the unlearned model should be \textit{different} from the pre-trained model in terms of the erasing concept in \textbf{UA} and should \textit{approach} the pre-trained model in \textbf{NRA} and \textbf{ORA}.} 
  \label{fig:finegrained_vis}
\end{figure}

\textbf{Fig.}\ref{fig:finegrained_vis} presents an illustrative comparison between SalUn and MiM‑MU across 3 fine-grained datasets. Generations by SalUn would loss background environment details (\textit{e.g.}, the 1st, 3rd, 4th images in UA column of Standford Dog,  the 4th image in ORA column of CUB-200) and exhibit blurred object edges (\textit{e.g.}, the 3rd, 4th, 5th images in UA column of Oxford Flowers) as well as overly smooth textures (\textit{e.g.}, the 3rd, 4th images in UA column of CUB-200,). In addition, many of SalUn’s retained-concept generations display over-saturation—colors look unnaturally vibrant artifact (\textit{e.g.}, 3rd image in ORA column of CUB-200). These suggest that SalUn’s erasure causes collateral degradation in image fidelity. By contrast, MiM-MU unlearned model generations exhibit more clear textures and natural colors, achieving higher fidelity.

\section{Limitations of Existing MU Methods}

In this section, we examine the limitations of existing concept erasure approaches to highlight some advantages of Mim-MU. We first investigate the resilience of unlearned models produced by SalUn, SDD, and Mim-MU when subjected to subsequent fine-tuning, revealing their vulnerability to concept resurgence. We then empirically reveal the failure of SalUn in handling multi-concept unlearning scenarios and the performance breakdown of SDD during its self-distillation process.

\textbf{Unlearning resilience of Mim-MU to subsequent fine-tuning.} \citet{suriyakumar2024unstable} reveals that subsequent fine-tuning on the unlearned model with even seemingly unrelated data, can inadvertently cause the model to relearn or resurge the previously erased concepts. Such vulnerability underscores the fragility of existing unlearning methods, \textit{i.e.}, the unlearned model would become unsafe again after further updates. To investigate the resilience of Mim-MU to such fine-tuning, we perform similar experiments for MiM-MU, SDD, and SalUn in \textbf{Tab.}\ref{tab:ft_robustness}. We observed that MiM-MU exhibits minor signs of concept recovery after additional fine-tuning, while SalUn and SDD stages an obvious recovery, especially when fine-tuned with a random subset of other remaining data. 


\begin{table*}[htbp]
\centering
\caption{\footnotesize UA of ``Abstractionism Style'' when subsequently training (fine-tuning) the unlearned model on the remaining data. Epoch-$i$ refers to UA of the unlearned model after fine-tuning $i$ epochs. \textbf{Class-wise} refers to fine-tuning the unlearned model with ``Objects in Seed Images Style'' data, and \textbf{Random-subset} refers to fine-tuning the unlearned model with a random subset from UnlearnCanvas benchmark without the erasing concept. $\Delta$ UA is computed as UA(Epoch-0) - UA(Epoch-8), reflecting the degradation in erasure performance (\textit{i.e.}, recovery of the erasing concept) after fine-tuning.}

\label{tab:abstractionism_ua}
\resizebox{0.95\textwidth}{!}{%
\begin{tabular}{lccccccc}
\toprule
\textbf{Fine-tune Data} & \textbf{Method} & \textbf{Epoch-0} & \textbf{Epoch-2} & \textbf{Epoch-4} & \textbf{Epoch-6} & \textbf{Epoch-8} & \textbf{$\Delta$ UA $\downarrow$} \\
\midrule
\multirow{3}{*}{Class-wise}  & SalUn   & $89.00\%$ & $86.00\%$ & $86.00\%$ & $82.00\%$ & $82.00\%$ & \congrat{$+7.00\%$} \\
          & SDD     & $100.00\%$ & $78.00\%$ & $87.00\%$ & $84.00\%$ & $90.00\%$ & $+10.00\%$ \\
          & Mim-MU  & $97.00\%$  & $89.00\%$ & $87.00\%$ & $93.00\%$ & $90.00\%$ & \congrat{$+7.00\%$} \\
\midrule 
\multirow{3}{*}{Random-subset}& SalUn & $89.00\%$ & $11.00\%$ & $8.00\%$  & $7.00\%$ & $13.00\%$ & \warn{$+76.00\%$} \\
          & SDD    & $100.00\%$ & $42.00\%$ & $31.00\%$ & $1.00\%$  & $23.00\%$ & \warn{$+77.00\%$} \\
          & Mim-MU & $97.00\%$  & $90.00\%$ & $92.00\%$ & $86.00\%$ & $86.00\%$ & \congrat{$+11.00\%$} \\
\bottomrule[1pt]
\end{tabular}%
\label{tab:ft_robustness}
}
\end{table*}

\begin{wraptable}[6]{r}
{0.48\textwidth} 
\vspace{-5mm}
\centering
\caption{\footnotesize Performances of SalUn and Mim-MU on multi-concepts unlearning.}
\label{tab:simultaneous_mu}
\resizebox{0.45\textwidth}{!}{%
\begin{tabular}{lcccccc}
\toprule
\textbf{Method} & \textbf{UA$\uparrow$} & \textbf{IRA$\uparrow$} & \textbf{CRA$\uparrow$} & \textbf{RA$\uparrow$} & \textbf{Avg.Acc$\uparrow$} \\
\midrule
SalUn & \warn{$10.83\%$} & $98.56\%$ & $98.92\%$ & $98.74\%$ & $69.44\%$ \\
Mim-MU & $98.33\%$ & $69.44\%$ & $92.75\%$ & $81.10\%$ & \congrat{$86.17\%$} \\
\bottomrule[1pt]
\end{tabular}%
}
\end{wraptable} \textbf{Failure of SalUn in multi-concept scenario.} We conducted experiments to simultaneously erase the same 6 styles as in \textbf{Tab.}~\ref{tab:sequential_mu}  and present the results in \textbf{Tab.}~\ref{tab:simultaneous_mu}. We appreciate SalUn’s pioneering contribution in introducing the weight saliency map to unlearning, however, we identified two practical limitations of this mechanism in the multi-concept setting: \textbf{(i)} \textit{Incomplete forgetting}: Even unlearning with 30 epochs, UA of SalUn is only $10.83\%$, while that of Mim-MU is $98.33\%$, which indicates the failure of SalUn in simultaneous concept erasure. \textbf{(ii)} \textit{Escalating computational overhead}: SalUn requires computing the prediction loss of the remaining data and regularizes it (add as a regularization term) to preserve model utility. Consequently, the total time grows to roughly twice the number of forgetting data. Also, it requires computing the saliency mask with forgetting data. As the forgetting set expands, both mask computation and model fine‑tuning scale almost linearly, leading to a steep runtime increase. In contrast, Mim-MU neither requires mask computation nor utility compensation, exhibiting an obvious time advantage. 

As for the failure of SalUn in success removal when unlearning simultaneously, we would like to attribute to two possible reasons: 
\textbf{(1)  Gradient Direction Cancellation}. SalUn constructs a single weight‑saliency mask by aggregating gradients from all “forget” concepts. When those prompts represent heterogeneous concepts, their gradients often conflict and summation partially cancel. As a result, they remain frozen, leaving key parameters untouched during unlearning and resulting only in superficial erasure.
\textbf{(2) Fixed edit‑budget bottleneck}. With the increase of target concepts, the pool of highly effective parameters would increase. A fixed threshold of saliency mask limits the number of editable weights, causing each concept to receive fewer dedicated updates. In contrast, Mim-MU localises concepts through a mutual‑information mechanism, which does not rely on weight attribution sparsity to reduce interference, scaling more naturally in the multi‑concept setting.

\begin{wrapfigure}[14]{r}{0.38\textwidth}
\vspace*{-5mm}
    \centering
\includegraphics[width=0.38\textwidth]{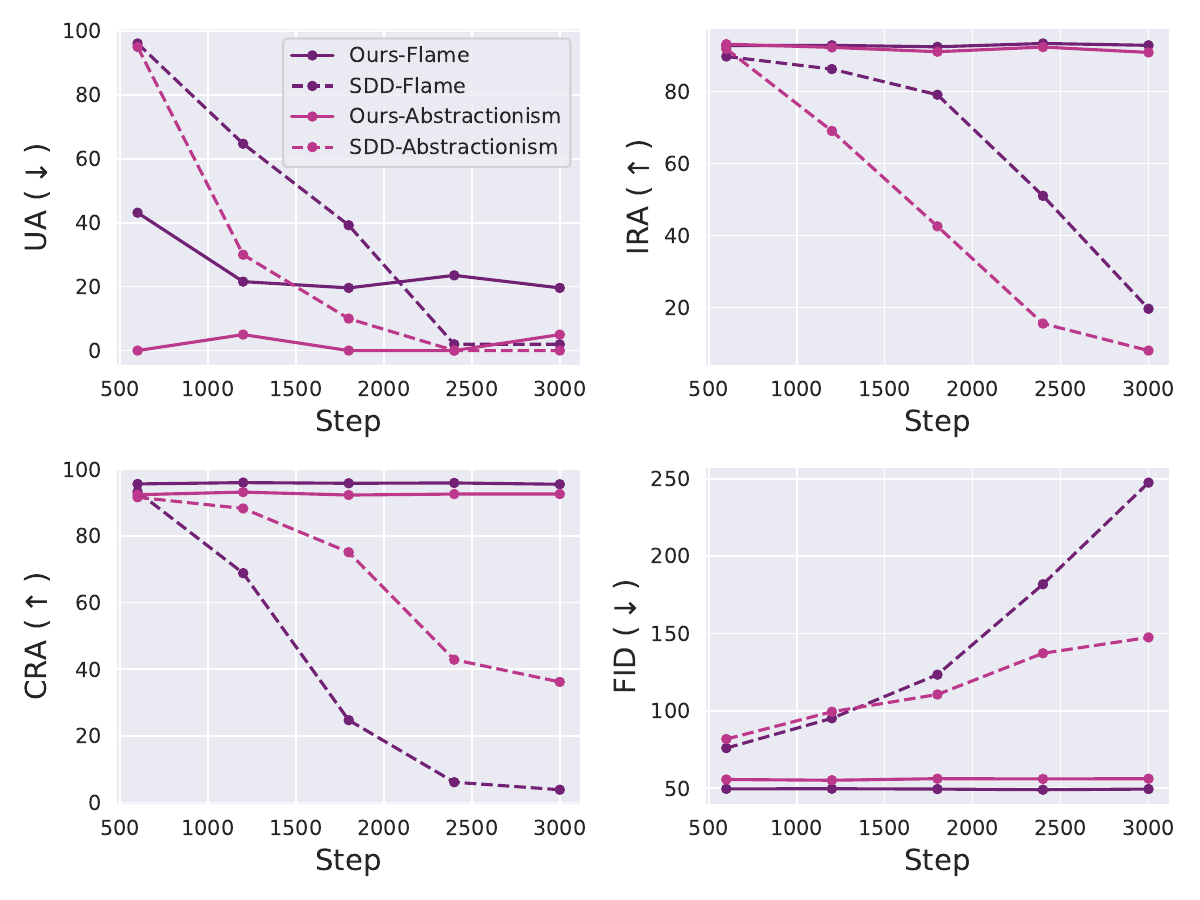}
\vspace*{-5mm}
    \caption{\footnotesize{
\textbf{Unlearning and retention during MiM-MU and SDD unlearning.} SDD exhibits a performance breakdown while ours well preserve the model utility as unlearning proceeds.}}
\label{fig:sdd_breakdown}
\end{wrapfigure}\textbf{Performance breakdown as self-distillation in SDD.} As elucidated in 
Sec.~\ref{sec:least_deviation}, the distillation teacher in SDD, which is the unconditional distribution of the unlearned model, will gradually deviate from the pre-trained model as unlearning proceeds. We empirically validate that continually distilling from it will induce model performance breakdown by unlearning ``Flame'' object and ``Abstractionism'' style in \textbf{Fig.}~\ref{fig:sdd_breakdown}. As  unlearning progresses, UA of MiM-MU and SDD both exhibit a progressive decline. However, the IRA and CRA of SDD experience a marked and rapid decline, while ours consistently converges to a high retainability. This highlights the instability of SDD’s self-distillation, which progressively erodes innocent knowledge, while MiM-MU could accurately locate and remove the undesired knowledge by identifying the mutual information. 

\textbf{Summary.} In summary, Mim-MU consistently preserves erasure effectiveness across different settings. Collectively, these findings suggest that Mim-MU provides a more reliable and practically applicable solution for concept erasure compared with existing approaches.

\section{Discussion}
\textbf{Further improvement on fine-grained concept erasure.} Although Mim-MU could achieve a comparable retention in fine-grained erasure without explicitly compensations, we also realize that there exists room for further improvement in mitigating unintended damage to other entangled concepts. In deed, fine-grained erasure refers to two concepts $y_1$ and $y_2$ are not independent with high correlations. Formally, this manifests as $p(Y_1, Y_2) \neq p(Y_1) p(Y_2)$. We think this challenge might be further improved using tools from information theory. Specifically, DiffusionPID~\cite{dewan2024diffusion} disentangles semantic dependencies by decomposing mutual information into unique, shared, and synergistic components, allowing precise attribution of each word’s unique contribution to the generated content. As visualized in their work, this decomposition provides a highly accurate alignment between different information components and concept locations in the generated image. Also, such decomposition from text space is convenient because this does not require access to the remaining data to compensate explicitly. In general, the entanglement of concepts could be formalized as statistically non-independent in our theoretical framework, leaving a good starting point for future study.

\section{Conclusion}
\vspace*{-2mm}
In this paper, we demonstrate that the widely adopted post-remedial compensations in existing concept erasure methods are inherently limited and insufficient for large-scale generative models. We identify the concept-related knowledge from the information-theoretic perspective of the diffusion model and propose to diminish the mutual information between the textual concept and the image to erase a concept. Our proposed MiM-MU minimally impacts other generations during removal by demanding the least deviations of the unlearned model from the pre-trained model. Extensive experiments demonstrate that MiM-MU can successfully remove undesired generations meanwhile preserving the general model utility without any compensation.

\newpage
\clearpage

{\small
\bibliographystyle{unsrtnat}
\bibliography{refs/unlearning}
}

\newpage
\appendix

\onecolumn

\section*{Appendix}
\setcounter{section}{0}
\setcounter{figure}{0}
\setcounter{prop}{0}
\makeatletter 
\renewcommand{\thefigure}{A\arabic{figure}}
\renewcommand{\theHfigure}{A\arabic{figure}}
\renewcommand{\thetable}{A\arabic{table}}
\renewcommand{\theHtable}{A\arabic{table}}

\makeatother
\setcounter{table}{0}

\setcounter{algorithm}{0}
\renewcommand{\thealgorithm}{A\arabic{algorithm}}
\setcounter{equation}{0}
\renewcommand{\theequation}{A\arabic{equation}}

\section{Preliminaries}\label{sec:preliminary}
\textbf{Diffusion models.} Generative models learn to model the true data distribution $p(x)$ by marginalize it out the latent variables and maximizing the \textbf{E}vidence \textbf{L}ower \textbf{BO}und (ELBO). Diffusion model establishes hierarchical latent variables by pre-defining the latent encoder as a linear Gaussian model, constructing a forward process $\{x_1, \dots, x_T\}$. The latent variable at timestep $t$ is designed as $x_t=\sqrt{\bar{\alpha}_t} x_0+\sqrt{1-\bar{\alpha}_t} \epsilon_0$, with variance schedule $\bar{\alpha}_t>0$. The ELBO of diffusion model is:
\vspace*{-0.5mm}
{\small{\begin{equation}
    \log p_{\theta}(x_0)  \geq \mathbb{E}_{q\left(x_{1: T} \mid x_0\right)}\left[\log \frac{p\left(x_0, x_{1: T}\right)}{q\left(x_{1: T} \mid x_0\right)}\right].
    \label{eq:ELBO}
\end{equation}}}The training objective turns into matching the approximate denoising transition step $p_\theta (x_{t-1}|x_t)$ to ground-truth denoising transition step $q(x_{t-1}|x_t, x_0)$ as closely as possible. Through re-parameterization tricks, the training objective turns into: 
\vspace{-1mm}
{\small
\begin{equation}
   \theta^* =  \arg \operatorname*{min}\limits_{\theta}~\mathbb{E}_{x\sim p(x),\epsilon, t}\left[\left\|\hat{\epsilon}_{\boldsymbol{\theta}}(x_{t},t) - \epsilon \right\|_{2}^{2}\right].
\end{equation}}where $\epsilon\sim\mathcal{N}(0,\mathcal{I})$ and $t\sim\text{Uniform}(1,\dots, T)$.
This objective is equivalent to learn the score function  from the score-based perspective of diffusion models. Through Tweedie's formula, the score function for the smoothed density is proportional to the predicted noise for $x_t$, $\nabla_{x_t} \log p(x_t)=-\dfrac{\hat{\epsilon}_{\boldsymbol{\theta}}(x_t)}{\sqrt{1-\bar\alpha_t}}$. The sampling process of diffusion model initiates with a random noise $x_T\sim \mathcal{N}(0, 1)$ and updates it iteratively with the predicted score. 

\textbf{Text-to-image (T2I) diffusion models.} T2I diffusion models aim to control the semantics of generated data through textual conditioning information $y$. Therefore, it learns conditional probability distribution $p(x|c)=p(c|x)p(x)/p(c)$, where $c:=\mathcal{T}(y)$ is the textual embedding of textual input $y$ and $\mathcal{T}$ denotes the text encoder. To avoid training an external classifier to provide the guidance term $p(c|x)$, \cite{ho2022classifier} proposes Classifier-free Guidance (CFG) to elegantly control how much the learned conditional model cares about the conditioning information with a simple scaler $\gamma$:
{\small\begin{equation}
\nabla_x \log p(x_t|c) = \gamma \nabla_x \log p(x_t|c) + (1-\gamma) \nabla_x \log p(x_t).
\end{equation}}

\textbf{Re-targetting based concept erasure methods.} 
A common way of unlearning is to align the behavior on an \textit{erasing concept} with that of an \textit{anchor concept} by the pre-trained model $\theta_{\text{P}}$ to alter its semantics. The anchor concept should be distinct from the erasing one, but not that divergent to severely hurts model performance. However, altering the semantics unavoidably confuses model performance on non-malicious \textit{retain concepts}, so that additional maintenance should be applied to patch up the degradation. Mathematically, these methods follow the following framework:
\vspace*{-1mm}
{\small{\begin{equation}
\begin{aligned}
\min_\theta\quad\underbrace{\left\|\Phi_{\theta_U}(x^f_t| c_{\text{erase}})-\Phi_{\theta_P}(x^f_t| c_{\text {anchor}})\right\|_2^2}_{ \text{Unlearning Term} \mathcal{L}_{\text{U}}}+\underbrace{\left\|\Phi_{\theta_U}\left(x^r_t| c_{\text{retain}}\right)-\Phi_{\theta_P}\left(x^r_t| c_{\text {retain}}\right)\right\|_2^2}_{\text{Retaining Term} \mathcal{L}_{\text{R}}}
\end{aligned}
\end{equation}}}Different methods focus on manipulating different responses $\Phi_{\theta}(\cdot)$ of the diffusion models, mainly including three types:  (1) the input cross-attention text-embedding projection, $\Phi_{\theta}(x_t|c) = W^TC$; (2) the intermediate cross-attention maps, $\Phi_{\theta}(x_t|c) = \mathcal{A}_{\theta}(x_t|c)$; (3) the output predicted noise $\Phi_{\theta}(x_t|c)=\epsilon_{\theta}(x_t,t,c)$. In this paper, we focus on manipulating the output predicted noise due to its straightforward connection with the information-theoretic view of diffusion models.


\section{Information-Theoretic Diffusion Model}
\subsection{Exact Log-likelihood Estimation by the Pre-trained Diffusion Model}\label{sec:inform_theoretic}
The pre-trained diffusion model behaves as a noisy channel capable of denoising Gaussian noise perturbed samples, denoted as $\vz \equiv \sqrt{\sigma(\logsnr)} \vx + \sqrt{\sigma(-\logsnr)} \eps$, where $\eps \sim \mathcal N(0, \mathbb I)$ and  $\alpha$ represents log SNR. \citet{guo2005mutual} demonstrated that the \textit{information} implicit in this Gaussian noise channel, denoted as $I(X;X_\logsnr)$, is \emph{exactly} the mean square error for optimal signal reconstruction. Mathematically, such information and its corresponding point-wise generalization are expressed as :

{\small \begin{equation}
 \ds I(\vx; \vz) = \half \mmse(\snr), 
 \quad\ds \KL\left[{p(\vz|\vx)}||{p(\vz)}\right]  = \half \mmse(\vx, \snr).
 \label{eq:pimmse}
\end{equation}} 

where $p(\vz) = \int p(\vz|\vx) p(\vx) d\vx$ is the marginal output distribution of sample and $\mmse(\vx, \snr) \equiv \mathbb E_{p(\vz|\vx)} \big[ \norm{\vx - \xhat^*(\vz, \snr)} 
\big]$ is the pointwise MMSE.
This provides an interpretation of \emph{information-theoretic} quantities with the \emph{estimation of optimal denoisers}. 

The pre-trained diffusion model establishes a transition path between true data distribution and the standard Gaussian distribution. \citet{kong2023information} apply thermodynamic variational inference along this path to recover the log likelihood for the data distribution. They consider sending samples from either the data distribution $p(\vz)$ or a standard Gaussian $p_G(\vz)=\mathcal N(0, I)$ through the Gaussian noise channel. The  marginal output distribution with Gaussian input is $p_G(\vz) = \int p(\vz|\vx) p_G(\vx) d\vx$ and corresponding MMSE for this Gaussian channel is $\mmse_G(\snr)$. Then they define the point-wise gap function $f(\vx, \snr)$ as
{\small \begin{equation}
    f(\vx, \snr) \equiv \KL\left[{p(\vz|\vx)}||{p_G(\vz)}\right] - \KL\left[{p(\vz|\vx)}||{p(\vz)}\right].
\end{equation}}In the limit of zero SNR, we have $\lim_{\snr \rightarrow 0} f(\vx, \snr) = 0$. In the high SNR limit, \cite{kong2023information} prove that $\lim_{\snr \rightarrow \infty} f(\vx, \snr) = \log \frac{p(\vx)}{p_G(\vx)}.$ Combining this with \eqref{eq:pimmse}, the exact log likelihood is:
\vspace*{-2mm}
{\small\begin{align}\label{eq:density}
\displaystyle
-\log p(\vx) &= -\log p_G(\vx) - \int_{0}^{\infty} d\snr \ds f(\vx, \snr) \nonumber \\
&= { -\log p_G(\vx)} - \half \int_{0}^{\infty} d\snr \left( {\mmse_G(\vx, \snr)} - \mmse(\vx, \snr) \right) \nonumber  \\
  &= - \half \int_{0}^{\infty} \mathbb{E}_{p(\boldsymbol{\epsilon})}\left[\left\|\boldsymbol{\epsilon}-\hat{\boldsymbol{\epsilon}}_\alpha\left(\boldsymbol{x}_\alpha\right)\right\|^2\right] d \alpha+\text{const}.
\end{align}}This represents density in terms of a Gaussian density and a correction that measures how much better we can denoise the target distribution than we could with an optimal denoiser for Gaussian source data. Furthermore, the density estimation is optimal for noise reconstruction error at different noise levels.

\subsection{Non-negative Mutual Information}\label{sec:non_neg_mi}
By substituting  $p(x)$ and $p(x|)$ with \eqref{eq:uncond_density} and \eqref{eq:cond_density}, we have the mutual information $\mathcal{I}(\vx,y)$ estimated by the pre-trained diffusion model as follows:
{\small\begin{equation}
\begin{aligned}
\mathcal{I}(\vx,y)=\left[ \log p(\vx | y)-\log p(\vx) \right]=\left[\half  \int_{0}^{\infty} \mathbb{E}_{\eps}\left[\left\|\eps-\hat{\eps}_{\theta_P}\left(\vx_\alpha\right)\right\|_2^2-\left\|\eps-\hat{\eps}_{\theta_P}\left(\vx_\alpha | y\right)\right\|_2^2\right] d \alpha\right]
\end{aligned}
\label{eq:ori_mutual}
\end{equation}}However, above estimation is not lower bounded, which warns that model performance might break down if the optimization does not stop timely. By expanding and rearranging \eqref{eq:ori_mutual}, we can have the second term in \eqref{eq:nonneg_mutual} equals to zero, benefiting from the orthogonality principle \cite{kay1993fundamentals}:
\vspace*{-1mm}
{\small \begin{equation}
\begin{aligned}
\mathcal{I}(\vx,y) &={[\overbrace{\left[\half \int_{0}^{\infty} \mathbb{E}_{\eps}\left[\left\|\hat{\eps}_\alpha\left(\vx_\alpha\right)-\hat{\eps}_\alpha\left(\vx_\alpha | y\right)\right\|_2^2\right] d \alpha\right.}^{\mathbb{I}^{+}(\vx ; y)}]} \\
& +2 \mathbb{E}_{p(y)}[1 / 2 \int_{0}^{\infty} \underbrace{\mathbb{E}_{p(\vx | y), \eps}\left[\left(\hat{\eps}_\alpha\left(\vx_\alpha\right)-\hat{\eps}_\alpha\left(\vx_\alpha | y\right)\right) \cdot\left(\hat{\eps}_\alpha\left(\vx_\alpha|y\right)-\eps\right)\right]}_{\equiv \mathcal{O}} d \alpha]
\end{aligned}   
\label{eq:nonneg_mutual}
\end{equation}}Compared with \eqref{eq:ori_mutual}, there are two advantages of \eqref{eq:nonneg_mutual}: (1) \textit{Non-negativity}: It is non-negative, avoiding breakdown of model performance during minimization. (2) \textit{Low Variance}: It avoids sampling random Gaussian noises $\eps$, which will introduce variance to estimations and optimizations. Also, this non-negative expression of the mutual information has been indicated to be effective in attributing the semantics in the generated image to corresponding textual words in prompts~\cite{kong2023interpretable, dewan2024diffusion}, locating the emergence of specific semantics during generation.

\section{Theoretical Derivation}
\subsection{Gradient of Mutual Information Minimization}\label{sec:theory_mi_grad}
We denote the mutual information at timestep $t$ as $\mathcal{I}_t(\vx,y):=\half\left\|\hat{\eps}_{\theta_P}\left(\vx_t|y\right)-\hat{\eps}_{\theta_P}\left(\vx_t\right)\right\|_2^2$, and then minimizing the unlearning objective in \eqref{eq:forget} corresponds to minimize each $\mathcal{I}_t(\vx,y)$. In diffusion model, we have $\vx_t=\sqrt{\bar{\alpha}_t}\vx+\sqrt{1-\bar{\alpha}_t}\eps, \vx = \dfrac{\tilde \vx_t - \sqrt{1-\bar{\alpha}_t} \hat\epsilon_{\theta_{U}}(\tilde \vx_t|y)}{\sqrt{\bar{\alpha}_t}}.$ Therefore, the gradient of $\mathcal{I}_t(\vx,y)$ w.r.t. the unlearned model $\theta_U$ is as the following: 
{\small{\begin{equation}
\begin{aligned}
    \dfrac{\partial\mathcal{I}_t(\vx, y)}{\partial \theta_U} &= 
    \dfrac{\partial \mathcal{I}_t(\vx, y)}{\partial \hat{\epsilon}_{\theta_P}} 
    \cdot \dfrac{\partial \hat{\epsilon}_{\theta_P}}{\partial \vx_t} 
    \cdot \dfrac{\partial \vx_t}{\partial \vx} 
    \cdot \dfrac{\partial \vx}{\partial \theta_U} \\    
    &= \mathbb{E}_{\eps} \biggl[
    (\hat{\eps}_{\theta_P}(\vx_t | y) - \hat{\eps}_{\theta_P}(\vx_t)) \cdot 
    \biggl(\dfrac{\partial \hat{\eps}_{\theta_P}(\vx_t | y)}{\partial \vx_t} - 
    \dfrac{\partial \hat{\eps}_{\theta_P}(\vx_t)}{\partial \vx_t}\biggr) \cdot 
    \sqrt{\bar{\alpha}_t} \cdot 
    -\dfrac{\sqrt{1-\bar{\alpha}_t}}{\sqrt{\bar{\alpha}_t}} \cdot 
    \dfrac{\partial \hat{\epsilon}_{\theta_U}(\tilde{\vx}_t | y)}{\partial \theta_U}
    \biggr] \\
    &= \mathbb{E}_{\eps} \biggl[- 
    \sqrt{1-\bar{\alpha}_t} \cdot 
    (\hat{\eps}_{\theta_P}(\vx_t | y) - \hat{\eps}_{\theta_P}(\vx_t)) \cdot 
    \biggl(\dfrac{\partial \hat{\eps}_{\theta_P}(\vx_t | y)}{\partial \vx_t} - 
    \dfrac{\partial \hat{\eps}_{\theta_P}(\vx_t)}{\partial \vx_t}\biggr)\cdot 
    \dfrac{\partial \hat{\epsilon}_{\theta_U}(\tilde{\vx}_t | y)}{\partial \theta_U}
    \biggr]. \\
    &= \mathbb{E}_{\eps} \biggl[w(t) \cdot 
    \underbrace{(\hat{\eps}_{\theta_P}(\vx_t | y) - \hat{\eps}_{\theta_P}(\vx_t))}_{\text{Pre-trained CFG}} \cdot 
    \underbrace{\biggl(\dfrac{\partial \hat{\eps}_{\theta_P}(\vx_t | y)}{\partial \vx_t} - 
    \dfrac{\partial \hat{\eps}_{\theta_P}(\vx_t)}{\partial \vx_t}\biggr)}_{\text{Pre-trained U-Net Jacobian}} \cdot 
    \underbrace{\dfrac{\partial \hat{\epsilon}_{\theta_U}(\tilde{\vx}_t | y)}{\partial \theta_U}}_{\text{Unlearned U-Net Gradient}}
    \biggr].
\end{aligned}
\end{equation}}}The obtained gradient in \eqref{eq:mutual_t_grad} can be decomposed into 3 components: (1) CFG term of the pre-trained diffusion model, (2) U-Net Jacobian of the pre-trained diffusion model, and (3) U-Net gradient of the unlearned model.

\subsection{Gradient of KL Divergence Minimization}\label{sec:theory_kl_equivalence}
In this part, we elucidate the equivalence between the approximated mutual information minimization and KL divergence minimization in \eqref{eq:kl_distill}.
This could be verified by back-propogating \eqref{eq:kl_distill} to the unlearned model:
\vspace*{-1mm}
{\small\begin{equation}
\begin{aligned}
    \dfrac{\partial \mathcal{D}_{\theta_P}^{\mathrm{KL}}(\vx)}{\partial \theta_U} &=  \dfrac{\partial \mathcal{D}_{\theta_P}^{\mathrm{KL}}(\vx)}{\partial \vx_t} \cdot \dfrac{\partial \vx_t}{\partial \vx} \cdot \dfrac{\partial \vx}{\partial \theta_U} \\
    &= \mathbb{E}_\eps \left[ \left(\dfrac{\partial \log p(\vx_t)}{\partial \vx_t} -  \dfrac{\partial \log p(\vx_t|y)}{\partial \vx_t}\right)\cdot \dfrac{\partial \vx_t}{\partial \vx}\cdot\dfrac{\partial \vx}{\partial \theta_U} \right]\\
    &=  \mathbb{E}_\eps \left[- \dfrac{1}{\sqrt{1-\bar{\alpha}_t}}\cdot-( \hat\epsilon_{\theta_P}(\vx_t|y)-\hat\epsilon_{\theta_P}(\vx_t)) \cdot \sqrt{\bar{\alpha}_t} \cdot - \dfrac{\sqrt{1-\bar{\alpha}_t} }{\sqrt{\bar{\alpha}_t}}\dfrac{\partial\hat\epsilon_{\theta_{U}}(\tilde \vx_t|y) }{\partial \theta_U}\right] \\
    &=\mathbb{E}_\eps \left[- (\hat\epsilon_{\theta_P}(\vx_t|y)-\hat\epsilon_{\theta_P}(\vx_t))\cdot \dfrac{\partial\hat\epsilon_{\theta_{U}}(\tilde \vx_t|y) }{\partial \theta_U} \right].\\
    &= \dfrac{1}{\sqrt{1-\bar{\alpha}_t}} \cdot \dfrac{\partial\mathcal{I}_t(\vx, y)}{\partial \theta_U}
\end{aligned}
\end{equation}}
This is proportional to the gradient of the U-Net omitted mutual information minimization in \eqref{eq:kl_distill_grad} .

\subsection{Equivalence between KL Divergence Minimization and Noise Prediction Alignment}\label{sec:theory_mim_mu}
In this part, we elucidate the equivalence between KL divergence minimization and noise prediction alignment in \eqref{eq:MUMI_loss}:
{\small{
\begin{equation}
   \mathrm{KL}(q_{\theta_U}(\vx_t|y)\|q^*_{\theta_U}(\vx_t|y)) = \mathbb{E}_{\eps} \left[\left\| \hat\epsilon_{\theta_U}(\vx_t|y) -\hat\epsilon_{\theta_P}(\vx_t)\right\|_2^2\right]
\end{equation}}}This could be verified by back-propagating the LHS and RHS respectively to examine their gradient: 
{\small\begin{equation}
\begin{aligned}
 \dfrac{\partial \mathrm{KL}(q_{\theta_U}(\vx_t|y)\|q^*_{\theta_U}(\vx_t|y))}{\partial \theta_U} &= \mathbb{E}_\eps \left[ \left(\dfrac{\partial \log p(\vx_t|y)}{\partial \vx_t} -  \dfrac{\partial \log p(\vx_t)}{\partial \vx_t}\right)\cdot \dfrac{\partial \vx_t}{\partial \vx}\cdot\dfrac{\partial \vx}{\partial \theta_U} \right]\\
    &=  \mathbb{E}_\eps \left[- \dfrac{1}{\sqrt{1-\bar{\alpha}_t}}\cdot( \hat\epsilon_{\theta_P}(\vx_t|y)-\hat\epsilon_{\theta_P}(\vx_t)) \cdot \sqrt{\bar{\alpha}_t} \cdot - \dfrac{\sqrt{1-\bar{\alpha}_t} }{\sqrt{\bar{\alpha}_t}}\dfrac{\partial\hat\epsilon_{\theta_{U}}(\tilde \vx_t|y) }{\partial \theta_U}\right] \\
    &=\mathbb{E}_\eps \left[(\hat\epsilon_{\theta_P}(\vx_t|y)-\hat\epsilon_{\theta_P}(\vx_t))\cdot \dfrac{\partial\hat\epsilon_{\theta_{U}}(\tilde \vx_t|y) }{\partial \theta_U} \right].
\end{aligned}
\end{equation}}

The gradient of MSE (RHS) is :
{\small\begin{equation}
\begin{aligned}
  \dfrac{\partial \mathbb{E}_{\eps} \left[\left\| \hat\epsilon_{\theta_U}(\vx_t|y) -\hat\epsilon_{\theta_P}(\vx_t)\right\|_2^2\right]}{\partial \theta_U} &= \mathbb{E}_{\eps} \left[(\hat\epsilon_{\theta_U}(\vx_t|y) -\hat\epsilon_{\theta_P}(\vx_t))\cdot \dfrac{\partial \hat\epsilon_{\theta_U}(\vx_t|y)}{\partial \theta_U} \right]
\end{aligned}
\end{equation}} where $\hat\epsilon_{\theta_P}(\vx_t)$ is distillation teacher and detached from computation graph. Therefore, we have the equivalent minimization in \eqref{eq:MUMI_loss}.

\section{UnlearnCanvas Visualizations}
In this section, we visualize the generations after unlearning to demonstrate the qualitative performance of different concept erasing methods. 

\subsection{Style and Object Unlearn}\label{sec:unlearncanvas_vis}
\begin{figure}[htbp]
  \centering
  \includegraphics[width=0.98\linewidth]{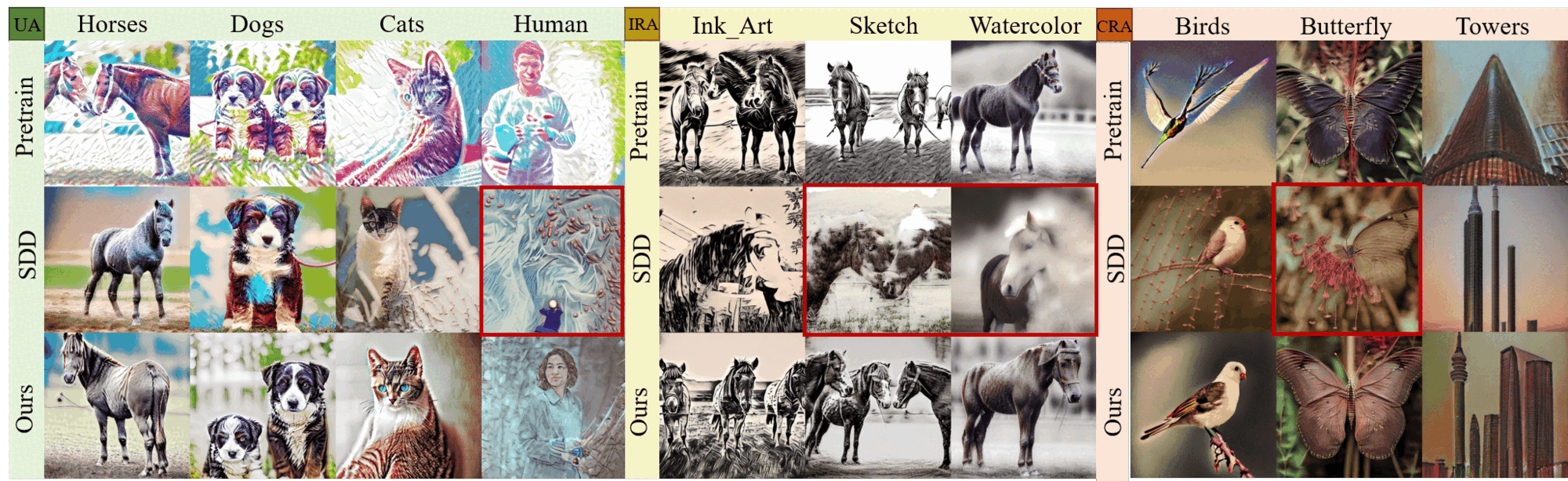}
  \caption{Visualization of the unlearning performance of MiM-MU and SDD on ``Cartoon'' style. The organization this figure follows \textbf{Fig.}~\ref{fig:UnlearnCanvas}. Both SDD and MiM-MU effectively erase  ``Cartoon'' style. SDD fails to preserve generating ``Sketch'' and ``Watercolor'' styles and ``Butterfly'' object, while ours demonstrates pretty retainability. }
\label{fig:uc_cartoon}
\end{figure}

\begin{figure}
    \centering
    \includegraphics[width=\linewidth]{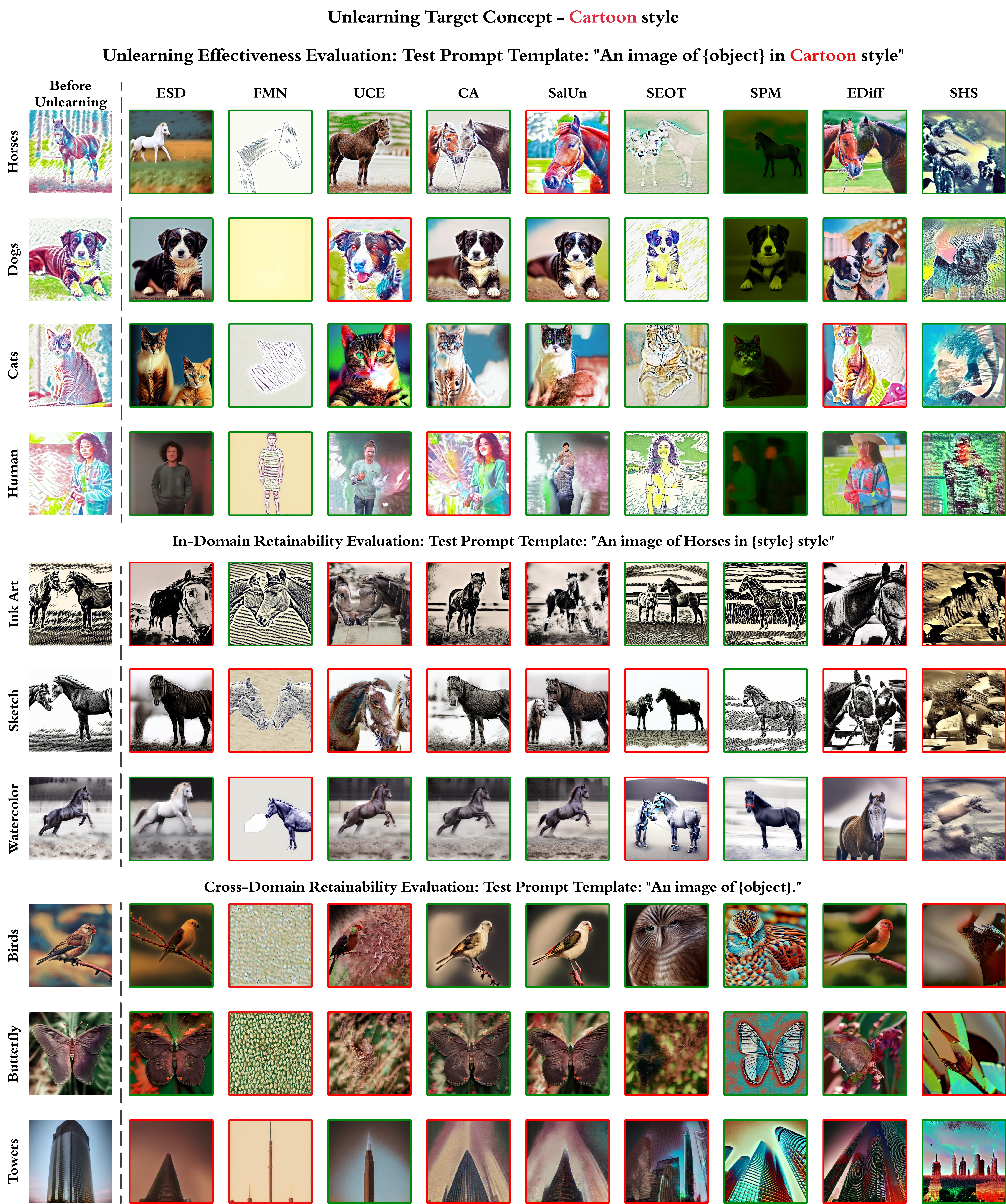}
    \caption{visualization of the unlearning performance of different methods on the task of style unlearning. Three text prompt templates are used to evaluate the unlearning effectiveness, in-domain retainability, and cross-domain retainability of each method. Images with \textcolor{green}{green} frame denote desirable results, while the ones with \textcolor{red}{red} frame denote unlearning or retaining failures.}
    \label{fig:ori_cartoon}
\end{figure}

\textbf{``Cartoon'' style unlearning performance comparisons} The original UnlearnCanvas benchmark provides abundant generation examples of all the 9 benchmarking methods in a case study of unlearning the ``Cartoon'' style. Both the successful and failure cases are
demonstrated in the context of unlearning effectiveness, in-domain retainability, and cross-domain
retainability. For the convenience of comparison, we copy it in \textbf{Fig.}~\ref{fig:ori_cartoon}. It is obvious that the original 9 benchmarking methods demonstrate less than satisfactory retainability, with evident generation failure and quality degradation of the remaining concepts. Furthermore, we visualize the generations of ``Cartoon'' unlearned model by SDD and MiM-MU in \textbf{Fig.}~\ref{fig:uc_cartoon}. Although SDD successfully prevents generating ``Cartoon'' style, it fails to generate ``Human'' when it is requested along with ``Cartoon'' style. Moreover, it fails to well generate ``Sketch'' and ``Watercolore'' style and ``Butterfly'' object. In contrast, MiM-MU effectively erases the undesired style and perfectly generates the other demanded concept. Meanwhile, it is able to generate the remaining concepts nearly as well as the pre-trained diffusion model. 

\begin{figure}[htbp]
  \centering
  \includegraphics[width=0.98\linewidth]{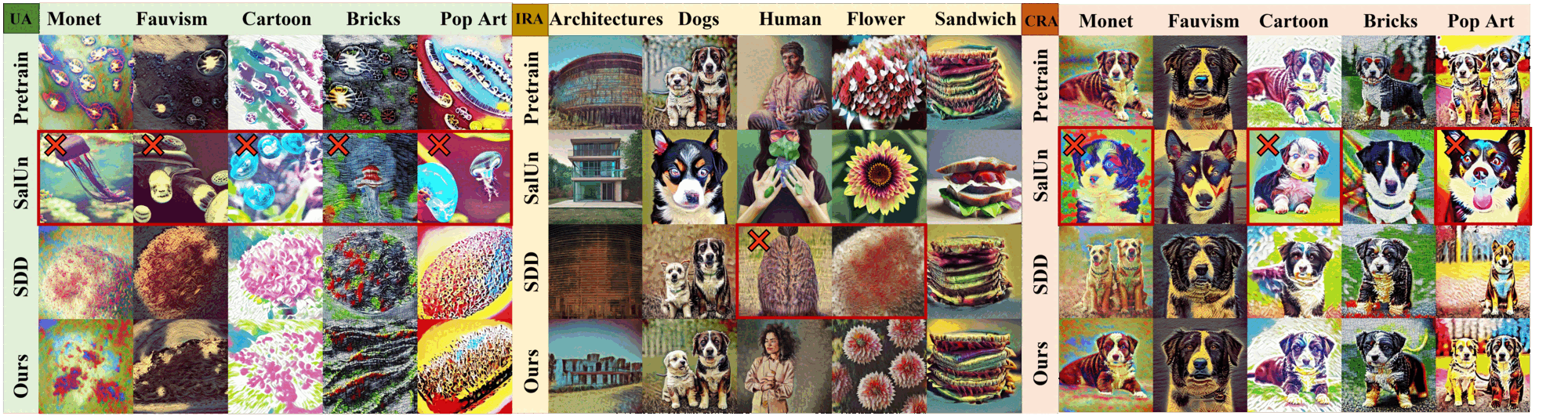}
  \caption{Visualization of the unlearning performance of MiM-MU and SDD on ``Jellyfish'' object. The organization this figure follows \textbf{Fig.}~\ref{fig:UnlearnCanvas}. SalUn fails to erase ``Jellyfish'' completely, and the painting styles of the other artists(\textit{e.g.}, ``Monet'', ``Cartoon'', and ``Pop Art'') exhibit obvious degradation when compared with the pre-trained diffusion model. In contrast, MiM-MU can effectively prevent generations of ``Jellyfish'' meanwhile preserving high-quality generations on other concepts.}
  \label{fig:jellyfish}
\end{figure}

\textbf{``Jellyfish'' object unlearning performance comparisons.}We provide visualizations of ``Jellyfish'' unlearned models by different unlearning methods in \textbf{Fig.}~\ref{fig:jellyfish}. The conclusion still holds true that existing methods fail to achieve satisfying retainability, while MiM-MU exhibits minimal degradation on model utility across various the remaining concepts.

\begin{figure}[htbp]
  \centering
  \includegraphics[width=0.98\linewidth]{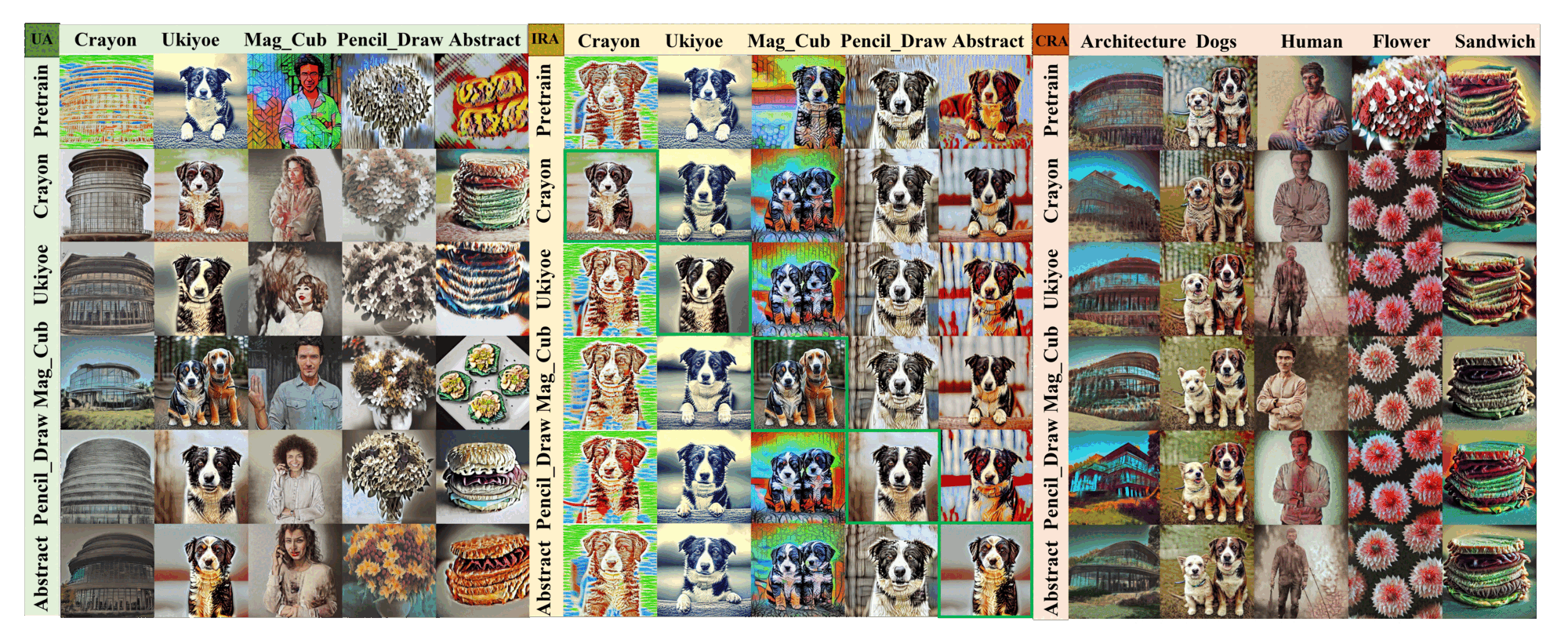}
  \caption{Visualization of the unlearning performance of different styles by MiM-MU. Each object of the pre-trained model in UA is combined with above 5 different styles successively to demonstrate the requested erasing style. From top to bottom, each row is the unlearning and retaining generations after erasing ``Crayon'', ``Ukiyoe'', ``Magic Cube'' (abbreviated as ``Mag Cub'' in figure), ``Pencil Drawing'' (``Pencil Draw''), ``Abstractionism'' (`Abstract''), respectively. The diagonal images (which are highlighted with green frameworks) in IRA indicate ``Dogs'' in current erasing style, therefore, it should not contain any painting style. }
  \label{fig:styles}
\vspace{-3mm}
\end{figure}

\textbf{5 Different artist styles unlearning by MiM-MU.}We demonstrate the generation quality after unlearning 5 different styles by MiM-MU in \textbf{Fig.}~\ref{fig:styles}. For each requested erasure, the MiM-MU unlearned model successfully removes the undesired painting style in UA.  In IRA and CRA, MiM-MU well preserve the original generative capability of the remaining styles and objects.  

\begin{figure}[htbp]
  \centering
  \includegraphics[width=0.98\linewidth]{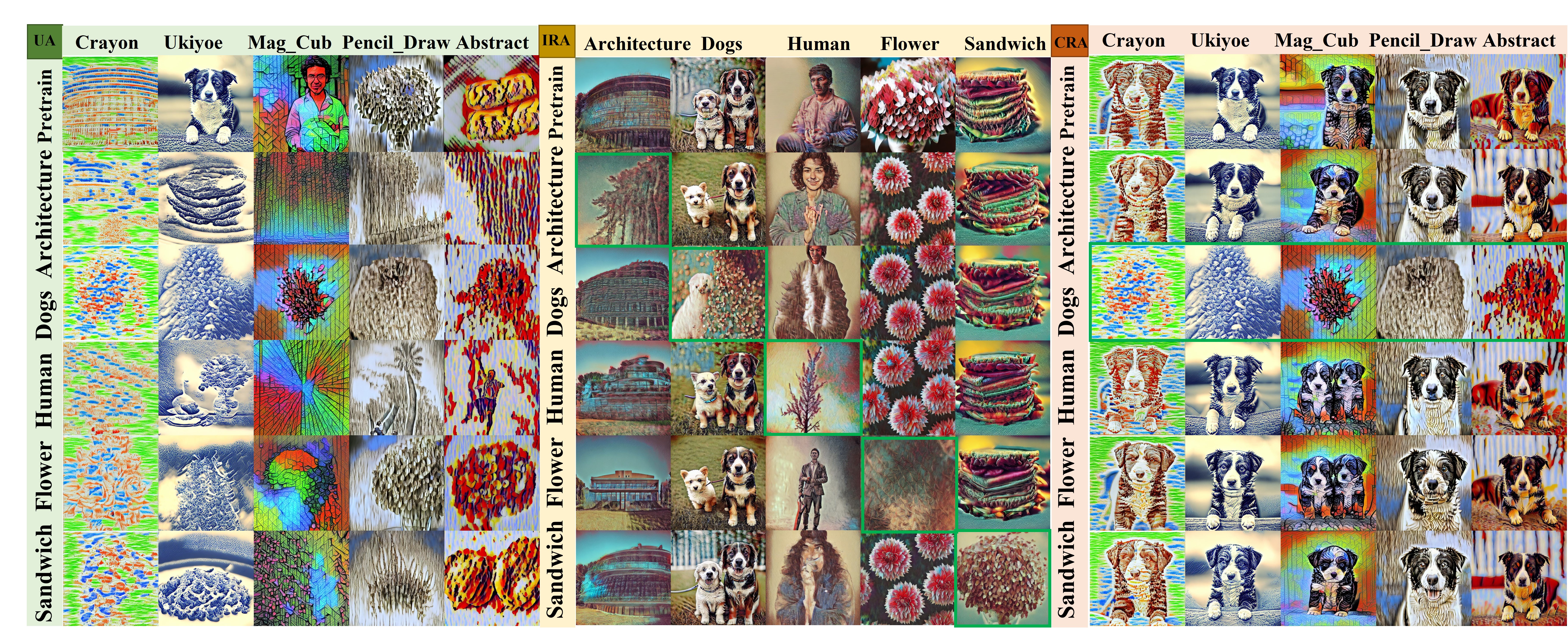}
  \caption{Visualization of the unlearning performance of different objects by MiM-MU. Each object of the pre-trained model in UA is combined with above 5 different styles successively to demonstrate the requested erasing style. From top to bottom, each row is the unlearning and retaining generations after erasing ``Architecture'', ``Dogs'', ``Human'' , ``Flower'', ``Sandwich'', respectively. The diagonal images (which are highlighted with green frameworks) in IRA indicate the erasing object in ``Seed Image'' style , therefore it should be empty if unlearning is successful. The third row of ``Dogs'' in CRA should be empty as well. }
  \label{fig:objects}
\vspace{-3mm}
\end{figure}

\textbf{5 Different objects unlearning by MiM-MU.}We demonstrate the generation quality after unlearning 5 different objects by MiM-MU in \textbf{Fig.}~\ref{fig:objects}. For each requested erasure, the MiM-MU unlearned model successfully removes the undesired objects in UA. In IRA and CRA, MiM-MU well preserve the original generative capability of the remaining styles and objects.  

\clearpage
\newpage
\subsection{Sequential Unlearning}\label{sec:sequential_unlearn}
\begin{figure}[htbp]
  \centering
  \includegraphics[width=0.9\linewidth]{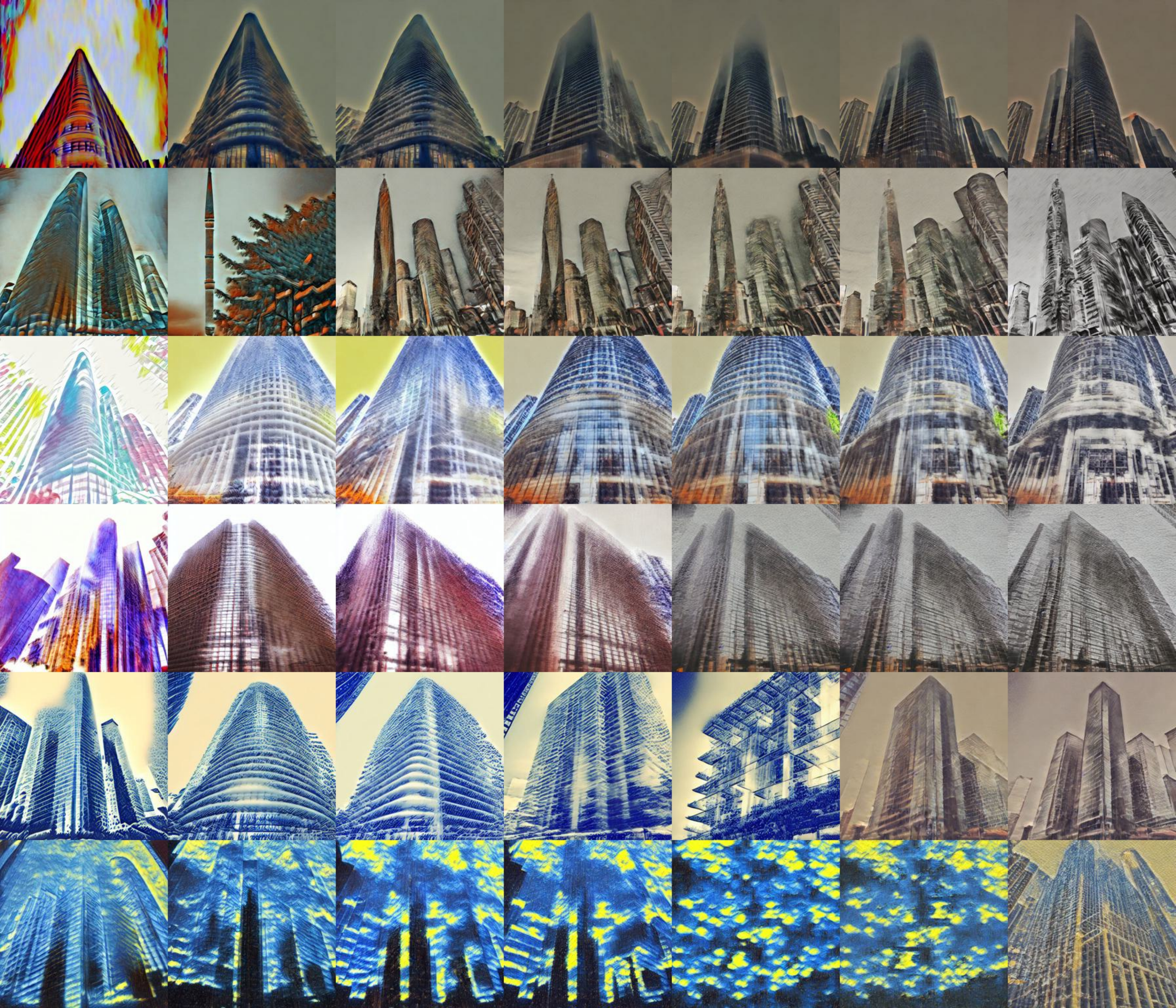}
  \caption{Visualizations of other In-domain concept generations during the sequential unlearning. From left to right, each column in sequence is the pre-trained model, followed by the models after unlearning ``Abstractionism'' ($\mathcal{T}_1$), ``Byzantine'' ($\mathcal{T}_2$), ``Cartoon'' ($\mathcal{T}_3$), ``Cold Warm'' ($\mathcal{T}_4$), ``Ukiyoe'' ($\mathcal{T}_5$), and ``Van Gogh'' ($\mathcal{T}_6$). From top to bottom, each row is the models' generation of ``A \textbf{Towers} in {\textbf{Artist}} style.'', where the \textbf{Artist} is the same sequence of the unlearning request. The results demonstrate that MiM-MU removes the generative capability for requested artist style thoroughly without any reemergence, indicating an exhaustive and reliable erasure operation, \textit{i.e.}, which is complete and does not spoil previous unlearning efforts. }
  \label{fig:sequential_UA}
\vspace*{-3mm}
\end{figure}

\begin{figure}[htbp]
  \centering
  \includegraphics[width=0.9\linewidth]{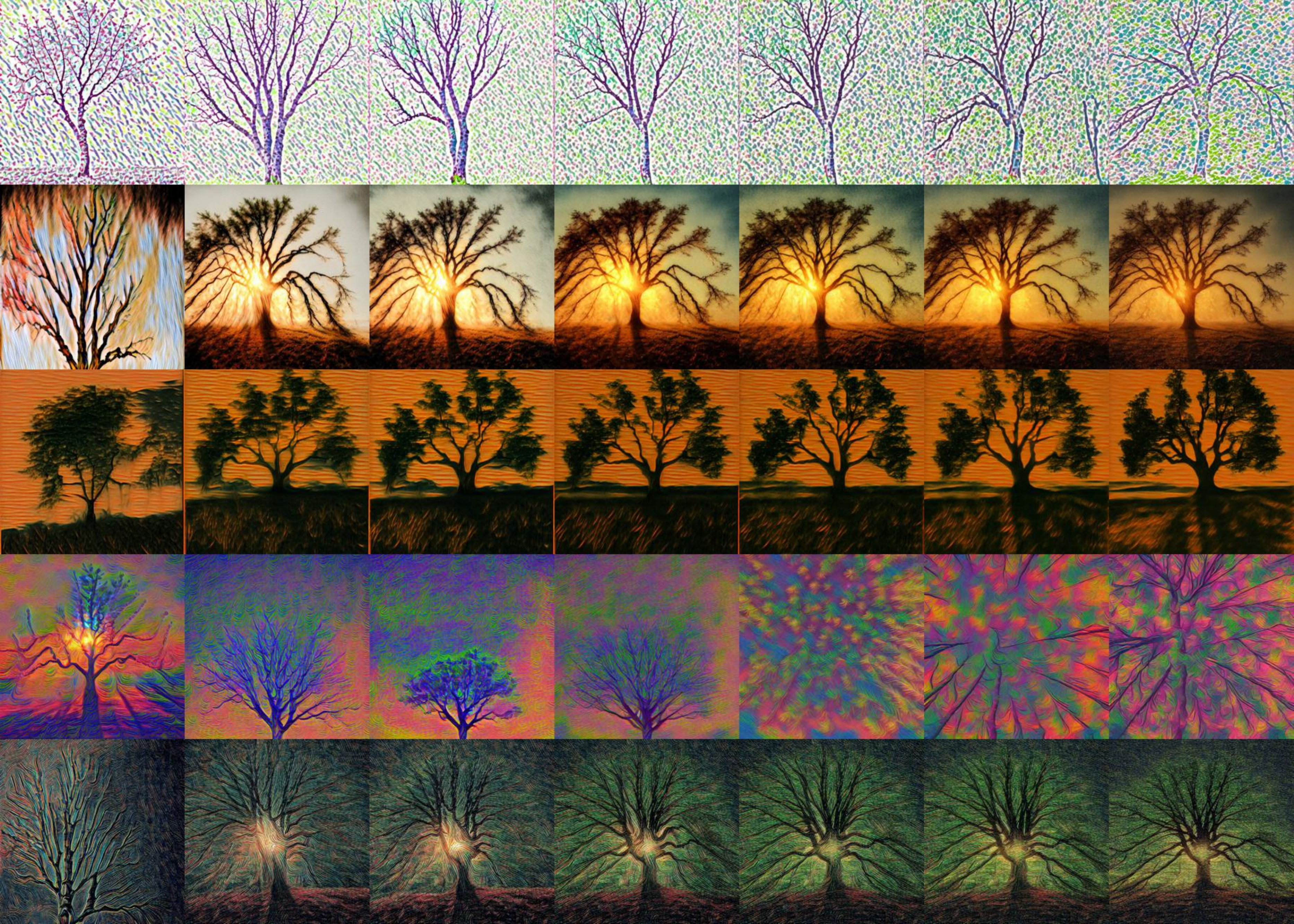}
  \caption{Visualizations of other In-domain concepts generations during the sequential unlearning. From left to right, each column in sequence is the pre-trained model, followed by the models after unlearning ``Abstractionism'' ( $\mathcal{T}_1$), ``Byzantine'' ($\mathcal{T}_2$), ``Cartoon'' ($\mathcal{T}_3$), ``Cold Warm'' ($\mathcal{T}_4$), ``Ukiyoe'' ($\mathcal{T}_5$), and ``Van Gogh'' ($\mathcal{T}_6$). From top to bottom, each row is the models' generation of ``A \textbf{Tree} in {\textbf{Artist}} style.'' where the \textbf{Artist} from top to bottom are ``\textbf{Dapple}'', ``\textbf{Warm Smear}'', ``\textbf{Glowing Sunset}'', ``\textbf{Color Fantasy}'', and ``\textbf{Neon Lines}'' respectively. The results demonstrate that our method can effectively preserve the generative capability for other artist styles during the sequential unlearning process.}
  \label{fig:sequential_IRA}
\vspace*{-3mm}
\end{figure}

\begin{figure}[H]
  \centering
  \includegraphics[width=0.9\linewidth]{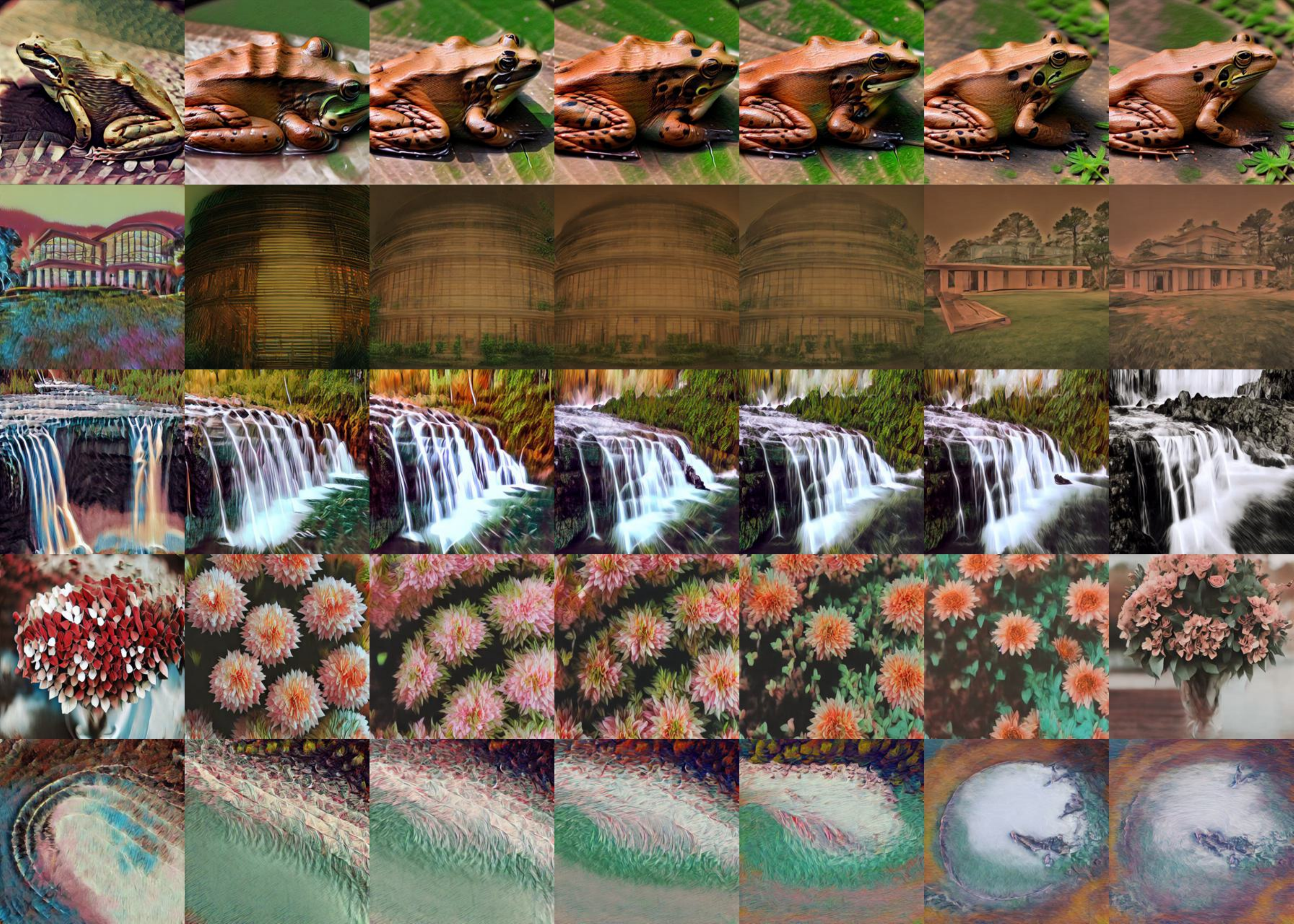}
  \caption{Visualizations of cross-domain concept (CRA) generations during the sequential unlearning process. From left to right, each column represents the pre-trained model, followed by the models after unlearning ``Abstractionism'' ($\mathcal{T}_1$), ``Byzantine'' 
 ( $\mathcal{T}_2$ ), ``Cartoon'' ( $\mathcal{T}_3$ ), ``Cold Warm'' ( $\mathcal{T}_4$ ), ``Ukiyoe'' ( $\mathcal{T}_5$ ), and ``Van Gogh'' ( $\mathcal{T}_6$ ). From top to bottom, each row depicts the models' generation of ``A {\textbf{Object}} in \textbf{Seed Images} style.'' where the \textbf{Object} from top to bottom are  ``\textbf{Frogs}'', ``\textbf{Architectures}'', ``\textbf{Waterfalls}'', ``\textbf{Flowers}'', and ``\textbf{Sea}'' respectively. The results indicate that our method successfully retains the generative ability for cross-domain concepts throughout the sequential unlearning process.}
  \label{fig:sequential_CRA}
\vspace*{-3mm}
\end{figure}

We visualize the unlearning and retaining performance of each unlearning request during sequential unlearning in \textbf{Fig.}~\ref{fig:sequential_UA}, \textbf{Fig.}~\ref{fig:sequential_IRA} and \textbf{Fig.}~\ref{fig:sequential_CRA}. The \textbf{first column} of each figure demonstrates generations of the \textbf{pre-trained} diffusion model. From the second column to the last column,  each column indicates generations of the \textbf{unlearned} model after $\mathcal{T}_1\sim\mathcal{T}_6$ unlearning request (``Abstractionism'' ($\mathcal{T}_1$), ``Byzantine'' ($\mathcal{T}_2$), ``Cartoon'' ($\mathcal{T}_3$), ``Cold Warm'' ($\mathcal{T}_4$), ``Ukiyoe'' ($\mathcal{T}_5$), and ``Van Gogh'' ($\mathcal{T}_6$)) respectively. In UA, the erasing style should be \textit{different} from the pre-trained diffusion model since the erasure is demanded. In IRA and CRA, the generation should be as similar as the pre-trained diffusion model throughout the sequential unlearning. 

\textbf{Fig.}~\ref{fig:sequential_UA} demonstrate the unlearning performance of each unlearning request. From top to bottom, each row stands for the style of each concept, \textit{e.g.}, \textbf{``Abstractionism'' , ``Byzantine'' , ``Cartoon'' , ``Cold Warm'', ``Ukiyoe'', and ``Van Gogh'' }.   In \textbf{Fig.}~\ref{fig:sequential_IRA}, from top to bottom, each row stands for a style of the remaining concepts, \textit{e.g.}, ``\textbf{Dapple}'', ``\textbf{Warm Smear}'', ``\textbf{Glowing Sunset}'', ``\textbf{Color Fantasy}'', and ``\textbf{Neon Lines}'' respectively. In \textbf{Fig.}~\ref{fig:sequential_CRA}, we demonstrate the cross-domain retainability with 6 objects, \textit{e.g.}, top to bottom are  ``\textbf{Frogs}'', ``\textbf{Architectures}'', ``\textbf{Waterfalls}'', ``\textbf{Flowers}'', and ``\textbf{Sea}''. As can be seen, MiM-MU can effectively remove the undesired style once the erasure is requested, and the erased style never reappears during subsequent unlearning requests. And MiM-MU is able to generate the remaining styles and objects with high quality, \textit{i.e.}, nearly identical to that of the pre-trained model. 

\section{Experiment Details}\label{sec:append_exp}
\subsection{Training Details}
For the forgetting dataset $\mathcal{X}_f$, we construct it by combining the erasing concept with each concept in the cross concept domain. For example, for unlearning ``Van Gogh'' style, we  combine it with 20 objects respectively, \textit{e.g.}, ``A Cat in Van Gogh style.''. For each combination, we use 3 images provided in the UnlearnCanvas dataset. For each unlearning request, the forget prompt $y$ is ``\{Artists\} Style'' for style unlearning (\textit{e.g.}, ``Van Gogh style'') and ``\{object\}'' for object unlearning (\textit{e.g.}, ``Dogs''). Then we fine-tune the cross-attention parameters of U-Net with our proposed loss and constructed $\mathcal{X}_f$ for 30 epochs, at a learning rate of $1\times 10^{-5}$ and batch size 1 for all the experiments. For evaluation, we evaluate with 5 seeds (188, 288, 388, 488 and 588) for all the experiments. The COCO-10k dataset used in this paper is downloaded from \url{https://github.com/OPTML-Group/AdvUnlearn}. All the experiments are conducted on RTX 4090 GPU. 

\subsection{Baselines}
In particular, we adopt the following training settings to reproduce baseline methods:
\begin{itemize}

\item SalUn~\cite{fan2023salun}: The code source of SalUn~\cite{fan2023salun} is \url{https://github.com/OPTML-Group/UnlearnCanvas}. We follow the implementation of SalUn in UnlearnCanvas benchmark. In our implementations, we run the weight saliency analysis with 1 epoch and unlearning stage with 10 epochs. The weight saliancy mask ratio is 0.5 and the learning rate is $1e-5$. 

\item SDD~\cite{kim2023towards}: The code source of SDD~\cite{kim2023towards} is \url{https://github.com/nannullna/safe-diffusion}. The forgetting concept for SDD is the corresponding name of artist (\textit{e.g.}, ``Van Gogh'') and object (\textit{e.g.}, ``Dogs''). We run SDD for 1k5 steps with a learning rate of $1e-5$, which is in line with the original implementation.
\end{itemize}


\subsection{Data Availability Statement}
The datasets used in this study are all publicly available and hosted on open-access repositories: 
\begin{enumerate}
    \item The UnlearnCanvas dataset is available at \url{https://huggingface.co/datasets/OPTML-Group/UnlearnCanvas}.
    \item Stanford Dogs dataset is available at \url{http://vision.stanford.edu/aditya86/ImageNetDogs/}.
    \item Oxford 102 Flowers is available at \url{https://www.robots.ox.ac.uk/~vgg/data/flowers/102/}.
    \item CUB-200 is available at \url{https://www.vision.caltech.edu/datasets/cub_200_2011/}.
    \item The COCO-1k subset used in this work is downloaded from the COCO-10k dataset, which is available at \url{https://github.com/OPTML-Group/AdvUnlearn.}.

\end{enumerate}

\section{Limitation}\label{sec:limitation}
In this paper, we focus on addressing the poor retainability of concept erasure in diffusion models. We propose a nuanced erasure, MiM-MU, to remove the knowledge of a concept from model parameters by minimizing the mutual information between the textual concept and semantic images. Our approach aims to effectively erase specific concepts while preserving the overall utility of the pre-trained model without requiring compensatory adjustments. However, several limitations remain open for future exploration. For computational feasibility, our mutual information minimization loss omits the UNet Jacobian term, which introduces approximations. These approximations may unintentionally weaken the ability to discriminate certain concepts during minimization, potentially impacting the unlearning effectiveness. Additionally, the whole sampling distribution of diffusion model is vast. In this paper, we focus on degrading the conditional sampling distribution of a specified concept. Future work should explore more accurate and efficient approximations and investigate how to comprehensively locate and degrade all the risky sampling distributions that are possible to generate an undesired concept. 


\clearpage





\end{document}